%% file: main.tex
\documentclass[10pt,twocolumn,letterpaper]{article}
\input{macros}
\usepackage[pagenumbers]{wacv} 

\usepackage{graphicx}
\usepackage{amsmath}
\usepackage{amssymb}
\usepackage{booktabs}
\usepackage[pagebackref,breaklinks,colorlinks]{hyperref}

\usepackage[capitalize]{cleveref}
\crefname{section}{Sec.}{Secs.}
\Crefname{section}{Section}{Sections}
\Crefname{table}{Table}{Tables}
\crefname{table}{Tab.}{Tabs.}

\def\wacvPaperID{1931} 
\def\confName{WACV}
\def\confYear{2024}
\usepackage{fancyhdr}
\fancypagestyle{firstpage}{%
  \fancyhf{}
  
  \fancyfoot[C]{\parbox{\textwidth}{\footnotesize\linespread{1}\selectfont\copyright\ IEEE 2023. Personal use of this material is permitted. Permission from IEEE must be obtained for all other uses, in any current or future media, including reprinting/republishing this material for advertising or promotional purposes, creating new collective works, for resale or redistribution to servers or lists, or reuse of any copyrighted component of this work in other works.}}
}

\begin{document}


\title{CCMR: High Resolution Optical Flow Estimation via\\ \underline{C}oarse-to-Fine \underline{C}ontext-Guided \underline{M}otion \underline{R}easoning}

\author{Azin Jahedi$^1$ \hspace{8mm} Maximilian Luz$^2$ \hspace{8mm} Marc Rivinius$^3$ \hspace{8mm} Andr\'es Bruhn$^1$\\
$^1$VIS, University of Stuttgart \hspace{6mm}$^2$RLL, University of Freiburg \hspace{6mm} $^3$SEC, University of Stuttgart\\
{\tt\small \{azin.jahedi,andres.bruhn\}@vis.uni-stuttgart.de}\\[-1.5mm]
{\tt\small luz@cs.uni-freiburg.de},
{\tt\small marc.rivinius@sec.uni-stuttgart.de}}



\maketitle

\begin{abstract} 
  Attention-based motion aggregation concepts have recently shown their usefulness in optical flow estimation, in particular when it comes to handling occluded regions. 
  However, due to their complexity, such concepts have been mainly restricted to coarse-resolution single-scale approaches that fail to provide the detailed outcome of high-resolution multi-scale networks.
  In this paper, we hence propose CCMR:
  a high-resolution coarse-to-fine approach that leverages attention-based motion grouping concepts to multi-scale optical flow estimation.
  CCMR relies on a hierarchical two-step attention-based context-motion grouping strategy that first computes global multi-scale context features 
  and then uses them to guide the actual motion grouping.
  As we iterate both steps over all coarse-to-fine scales, we adapt cross covariance image transformers to allow for an efficient realization while maintaining scale-dependent properties.
  Experiments and ablations demonstrate 
  that our efforts 
  of combining multi-scale and attention-based concepts
  pay off.  
  By providing 
  highly
  detailed flow fields 
  with strong improvements in both occluded and non-occluded regions, our CCMR approach not only outperforms both the corresponding single-scale attention-based and multi-scale attention-free baselines by up to 23.0\% and 21.6\%, respectively, it also achieves state-of-the-art results, ranking first on KITTI 2015 and second on MPI Sintel Clean and Final. 
  Code and trained models are available at  {\small\texttt{https://github.com/cv-stuttgart/CCMR}}.
\end{abstract}
\thispagestyle{firstpage}

\section{Introduction}
\label{sec:intro}
\begin{figure}[t]
    \centering
    \input{figures/teaser_new}
    \vspace{-2mm}
    \caption{Comparison of our method to the ground truth (first row), recent approaches from the literature (second row) and the 
    baselines (third row). Our method offers more structural details in the foreground and background (see boxes). 
    }
    \label{fig:teaser}
    \vspace{-2mm}
\end{figure}
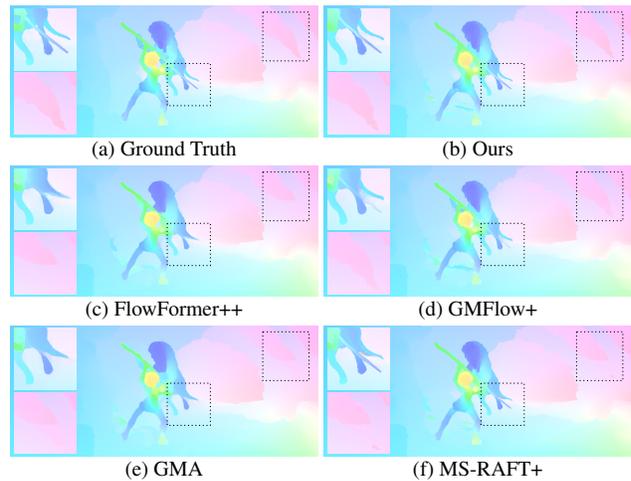
Estimating the optical flow in terms of the 2D displacement field between two consecutive frames of an image sequence is a fundamental task in computer vision.
It has found wide-spread application, ranging from autonomous driving and robotics over tracking and surveillance to frame interpolation and video compression.

Despite decades of research, the accurate estimation of optical flow 
still
remains challenging.
In this context,
most contemporary approaches rely on convolutional neural networks (CNNs) \cite{Dosovitskiy2015_FlowNet,Ilg2017_FlowNet2,Ranjan2017_SPyNet,sunPWCNetCNNsOptical2018,teedRAFTRecurrentAllPairs2020}. However, they lack global context, making reasoning about occlusions or large displacements difficult.
Being capable of modeling dynamic long-range interactions, transformers and other attention-based methods go far beyond the fixed receptive field of classical CNNs. 
Hence, it is not surprising  that by providing this otherwise missing global information, they showed great potential in various computer vision tasks and advanced the field of optical flow estimation with respect to the aforementioned shortcomings
\cite{jiangLearningEstimateHidden2021,zhangSeparableFlowLearning2021,huangFlowFormer2022,xuGMFlowLearningOptical2022,zhaoGlobalMatchingOverlapping2022,shiFlowFormerplusplus2023,xuUnifyingFlowStereo2022}.
Particularly interesting in this context is GMA \cite{jiangLearningEstimateHidden2021}, which has shown improvements in occluded areas by attention-based aggregation of motion features and features of the reference frame. 
Recently, this 
approach
has also been used by several other methods \cite{suiCRAFT2022,jungAnyFlow2023,dongMatchFlowGMA2023} as an improved backbone.  


A second somewhat orthogonal successful concept, that seeks to embed the estimation in a more global context,
is the computation of optical flow in a multi-scale coarse-to-fine setting. It dates back to classical methods
\cite{Brox2004,anandanComputationalFrameworkAlgorithm1989} and early CNN-based approaches \cite{Ranjan2017_SPyNet,sunPWCNetCNNsOptical2018,hurIterativeResidualRefinement2019} where it primarily helped to improve the estimation of large displacements. In current methods, however, multi-scale approaches are hardly employed. Instead, they mainly follow the architectural shift introduced by RAFT \cite{teedRAFTRecurrentAllPairs2020} by estimating the optical flow on a single coarse scale and then upsampling the final result \cite{jiangLearningEstimateHidden2021,zhangSeparableFlowLearning2021,luoLearningOpticalFlow2022,Sun2022_NeurIPS,huangFlowFormer2022,shiFlowFormerplusplus2023}. Recently, however, MS-RAFT(+) \cite{jahediMultiScaleRAFTCombining2022,jahediMSRAFTplus2022} has shown that such schemes are still relevant and can,
compared to single-scale approaches, not only improve robustness regarding ambiguous matches, but also produce more detailed flow fields. Thereby, valuable initializations from coarser scales allow them to operate more effectively on higher resolutions.


Evidently, it seems desirable to combine the strength of both attention-based motion guidance and high-resolution multi-scale schemes by integrating both concepts in a single framework.
However, naively applying the (token) attention mechanism 
to the high resolution finer scales is prohibitive due to its quad\-ratic complexity 
in the number of tokens (w.r.t.\ time and memory),
thus requiring alternatives. 
Moreover, the desire to obtain a global context makes any of the local-attention based approaches \cite{ramachandranStandAloneSelfAttentionVision2019,vaswaniScalingLocalSelfAttention2021} proposed specifically to address the complexity problem less suitable.
Methods
attempting to retain global reasoning via hierarchical structures (e.g., \cite{liu_2021_swin,wang_2021_pyramid-vision-transformer,fanMultiscaleVisionTransformers2021,li_2022_mvitv2}; generally local at a scale) 
do inherently not retain their input resolution. 
This makes them less desirable, as our approach requires scale-specific global representations in all the coarse-to-fine scales.

An attractive option in this context seem to be cross-covariance image transformers (XCiT) \cite{xcit_2021_ali}, which propose cross-covariance attention (XCA) as an alternative to token attention \cite{vaswani_2017_attention} by essentially transposing feature-channels and tokens.
This leads to an attention matrix quadratic over the number of input channels instead of tokens, making its complexity linear regarding the input size. 
Even though XCA loses the direct spatial structure inherent in standard token-attention, it retains implicit global interactions through the attention matrix, which can be seen as an expression of global channel statistics.
As a result, XCiT 
performs 
favorably against various transformer-based models, even outperforming some \cite{wang_2021_pyramid-vision-transformer, liu_2021_swin, zhang_2021_multiscale_longformer}.

\medskip
\noindent {\bf Contributions.}
We propose CCMR, an optical flow approach that combines the benefits of attention-based motion aggregation concepts and high-resolution multi-scale approaches. In this context, our contributions are fourfold: (i) First, we consistently integrate context-based motion grouping concepts into a high-resolution coarse-to-fine estimation framework. This allows us to obtained detailed flow fields that also offer a high accuracy in occluded regions (see \cref{fig:teaser}).
(ii) Secondly, in this context, we propose a two-step motion grouping strategy, where we first compute global self-attentive context features which are then used to guide the motion features iteratively over all scales. Thereby, an XCiT-based context-guided motion reasoning makes the processing over all coarse-to-fine scales feasible, while refraining from explicit
patchification enables us to maintain scale-specific content (in contrast to\cite{xcit_2021_ali}).
(iii)~Moreover, revisiting the U-Net style feature extractor of our multi-scale backbone, we increase its expressiveness while, independently, we reduce the size of the entire optical flow model by 28\% using a leaner context encoder. Although these architectural changes are small,
they strongly contribute to the improvements in non-occluded regions. 
(iv)~Finally, experiments and ablations demonstrate the strong performance of our approach and the benefits of the underlying concepts. It achieves state-of-the-art results ranking first on KITTI 2015 and second on MPI-Sintel, Clean and Final. 

\section{Related Work}
\label{related-work}

\noindent\textbf{Multi-Scale Optical Flow Estimation.}
Most early neural-network based optical flow methods \cite{Ranjan2017_SPyNet,sunPWCNetCNNsOptical2018,hurIterativeResidualRefinement2019,hofingerImprovingOpticalFlow2020}
owe their success to multi-scale estimation 
using coarse-to-fine warping schemes.
These schemes estimate residual motion at each scale by iteratively proceeding from coarse to fine levels, compensating for previously estimated motion by warping the second frame towards the first, via the current flow estimate.
First explored in classical approaches \cite{anandanComputationalFrameworkAlgorithm1989,Brox2004}, and therein found to be beneficial for estimating large motion and avoiding local minima, 
they have proven to be similarly successful for neural networks
\cite{sunPWCNetCNNsOptical2018,huiLiteFlowNetLightweightConvolutional2018}.
Subsequent works 
focused on improving iterative refinement via weight sharing \cite{hurIterativeResidualRefinement2019} and replacement of warping with sampling \cite{hofingerImprovingOpticalFlow2020}, avoiding ambiguities through ghosting effects. 

Introducing a paradigm shift, RAFT \cite{teedRAFTRecurrentAllPairs2020} suggested the single-scale estimation of flow via a recurrent refinement strategy, now found in many contemporary and state-of-the-art approaches \cite{jiangLearningEstimateHidden2021,jahediMultiScaleRAFTCombining2022,zhangSeparableFlowLearning2021,huangFlowFormer2022,luoLearningOpticalFlow2022,dongMatchFlowGMA2023,xuUnifyingFlowStereo2022}.
In each step, matching-costs are sampled based on the current flow estimate from a cost volume and fed through a recurrent unit to compute a residual flow update. 
While this poses a break from the until then predominant coarse-to-fine methods towards single-scale estimation, 
newer approaches, including DIP \cite{zhengDIPDeepInverse2022} and GMFlow(+) \cite{xuGMFlowLearningOptical2022,xuUnifyingFlowStereo2022} found it again beneficial to include a refinement stage on a secondary, finer scale, somewhat resembling earlier coarse-to-fine models.
Recently, AnyFlow \cite{jungAnyFlow2023} proposes to aggregate the recurrent update with finer scale image features estimating a more detailed flow, however not in a coarse-to-fine manner.

Recently, MS-RAFT(+) \cite{jahediMultiScaleRAFTCombining2022,jahediMSRAFTplus2022}, the architectural base of our method, reshaped RAFT into a coarse-to-fine estimation scheme using weight-sharing across scales, while relying on both multi-scale features and explicit multi-scale supervision.
Thereby, the use of a coarse-to-fine 
scheme allows 
the estimation of flow at a higher final resolution, 
improving accuracy and preserving finer details, while also retaining the capability for estimating large motion via coarser scales.

In addition to the benefits inherited from MS-RAFT(+), our approach utilizes motion grouping concepts based on global self-attentive context and more expressive 
encoders, explicitly addressing occlusion handling and matching quality, respectively.
In contrast to GMFlow(+) \cite{xuGMFlowLearningOptical2022,xuUnifyingFlowStereo2022} and DIP \cite{zhengDIPDeepInverse2022} that rely on additional refinement steps on a second scale, 
we consider   
a consistent coarse-to-fine method using four scales,
operating at a higher final resolution ($\frac{1}{2} \times (h, w)$ vs.\ $\frac{1}{4} \times (h, w)$). 

\medskip
\noindent\textbf{Occlusions and Ambiguities.} 
Several approaches aim to explicitly predict occlusion masks, either in a supervised or self-supervised fashion \cite{hurIterativeResidualRefinement2019, neoralContinualOcclusionOptical2019,ilgOcclusionsMotionDepth2018,jeongImposingConsistencyOptical2022} or indirectly in an unsupervised way \cite{Zhao2020_MaskFlownet}.
We, however, focus on enabling networks to deal with occlusions and ambiguities intrinsically, intending to create a more holistic motion reasoning process without the need for any special training procedures. 

In this context, both GMA \cite{jiangLearningEstimateHidden2021} and KPA-Flow \cite{luoLearningOpticalFlow2022} aggregate motion-features based on the matching costs and current flow estimate with context features from the reference image via a token cross-attention mechanism.
While GMA aims to resolve ambiguities and occlusions 
by incorporating global context-motion information,
KPA-Flow takes inspiration from the smoothness-terms of classical energy-based optical flow approaches and aggregates motions only locally. 
Recently, GMA motion aggregation modules were also directly adopted by other methods, such as AnyFlow \cite{jungAnyFlow2023}, MatchFlow \cite{dongMatchFlowGMA2023} and CRAFT\cite{suiCRAFT2022}, as an improved backbone compared to RAFT.

GMFlow(+) \cite{xuGMFlowLearningOptical2022,xuUnifyingFlowStereo2022}, on the other hand, follows the motion aggregation ideas  
used by GMA \cite{jiangLearningEstimateHidden2021} and KPA-Flow \cite{luoLearningOpticalFlow2022} by performing an attention-based refinement step using self-similarity of image-features to propagate flow (instead of motion-features) into occluded and ambiguous areas.
SeparableFlow \cite{zhangSeparableFlowLearning2021}, 
instead, 
aims to resolve difficult matches by applying an aggregation strategy to the cost volume itself.
Finally, SKFlow \cite{Sun2022_NeurIPS} 
seeks
to improve over GMA and RAFT by enlarging the receptive field of the convolutional layers in the motion encoder and update blocks specifically via their introduced super kernels.

In this work, we leverage motion grouping concepts to handle occlusions in a high-resolution coarse-to-fine framework. 
In contrast to GMA, our method not only operates on multiple scales, but it also relies on a two-step motion reasoning strategy that uses global self-attentive context features to guide the motion features.  
Thereby, the use of our motion reasoning concepts 
across all coarse-to-fine scales up to the finest scale ($\frac{h}{2}, \frac{w}{2}$) is enabled by exploiting XCiT which has linear complexity over the number of pixels.
 Moreover, unlike other approaches which use XCiT in other vision tasks \cite{phan_2022_xcit_patch,maaz_2023_edgenext,panteleris_2022_pe-former,zamirRestormerEfficientTransformer2022}, we refrain from the explicit patchification of its inputs originally proposed by XCiT and apply it directly to features at different scales maintaining their scale-specific properties. 
Besides, we further extend XCiT application beyond 
its original work \cite{xcit_2021_ali}, employing it not only for self- but also for cross-attention.

\begin{figure}
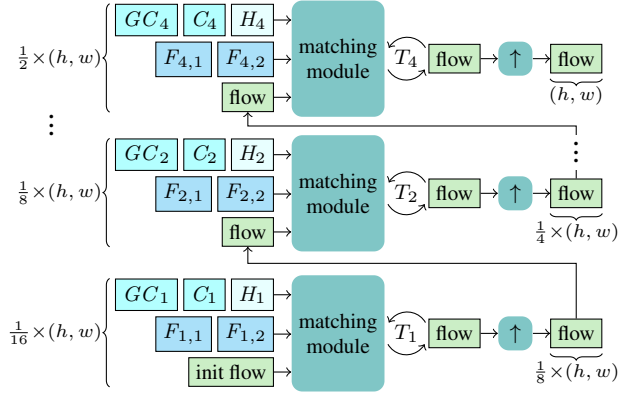

    \centering
    \include{figures/multi-scale}
    \vspace{-9mm}
    \caption{Coarse-to-fine architecture. The flow is first estimated on the coarsest scale, upsampled and used as initialization on the next finer scale. 
    Note that the third scale is omitted for compactness.}
    \label{fig:coarse_to_fine}
    \vspace{-1mm}
\end{figure}

\section{Approach}
\label{approach}

\subsection{Multi-Scale Flow Estimation}
\label{subsec:approach:MS-estimation}
Our CCMR approach estimates the optical flow in a coarse-to-fine manner via recurrent updates, using a shared Gated Recurrent Unit (GRU), 
based on MS-RAFT+ 
\cite{jahediMSRAFTplus2022}. Before starting with the coarse-to-fine estimation, for each scale $s$, image features $F_{s,1}, F_{s,2}$ of both frames $I_1, I_2$ are computed for matching. Moreover, context features $C_s$ and, based on that, global context features $GC_s$, and the current scale's initial hidden state $H_s$ of the recurrent unit are computed from the reference frame $I_1$. 
\cref{fig:coarse_to_fine} illustrates the coarse-to-fine estimation process. Starting from the coarsest scale at $\frac{1}{16} \times (h, w)$, where $h$ and $w$ denotes the height and width of $I_1$, the flow is computed based on the above-mentioned features $F_{1,1}, F_{1,2}, C_1, GC_1, H_1$, as well as an initial flow which is zero. After $T_1$ recurrent flow updates, the estimated flow is upsampled using a shared $\times 2$ convex upsampler, where the flow serves as initialization for the matching process on the next finer scale. This procedure continues until the flow is computed on the finest scale at $\frac{1}{2} \times (h, w)$ and upsampled 
to the original resolution. For supervision, we use the same multi-scale multi-iteration loss as in \cite{jahediMultiScaleRAFTCombining2022}. To this end, the upsampled flow in each 
scale and each iteration 
is bilinearly upsampled to the original resolution for computing its difference to the ground truth.

In the following we detail on the key components of our CCMR architecture: the improved multi-scale feature consolidation unit to extract $F_{s,1}, F_{s,2}, C_s$ and $H_s$, the computation of the global self-attentive context $GC_s$, the use of the global context for attention-based motion reasoning, and the efficient realization of both components via XCiT-based self- and cross-attention, enabling the integration of such concepts
in our high-resolution coarse-to-fine scheme.

\begin{figure}
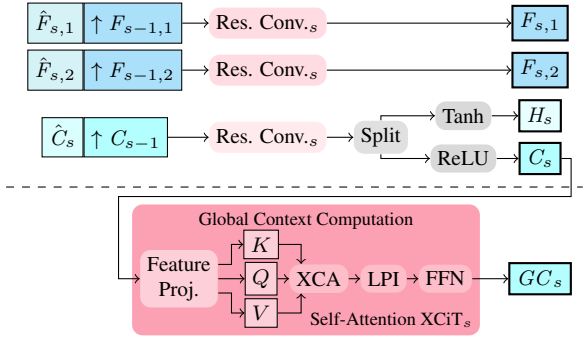

    \centering
    \include{figures/context-feature-transform}
    \vspace{-9mm}
    \caption{Improved multi-scale feature consolidation (top) and global context computation (bottom). Outputs shown on the right 
    are used as inputs in the matching module in \cref{fig:coarse_to_fine}. 
    }
    \label{fig:consolidation-GC}
    \vspace{-1mm}
\end{figure}

\subsection{Improved Mutli-Scale Feature Consolidation}
\label{subsec:feat_consolidation}

MS-RAFT(+)\cite{jahediMultiScaleRAFTCombining2022,jahediMSRAFTplus2022} proposed to extract multi-scale image and context features
using a U-Net-style feature extractor, 
inspired by feature pyramid networks \cite{linFeaturePyramidNetworks2017}.
To this end,
intermediate features are computed in a top-down manner and then, to obtain multi-scale features,
the better-structured finer features
$\hat{F}_{s,1}, \hat{F}_{s,2}$ and $\hat{C}_{s}$
are semantically boosted by 
consolidating them with the deeper coarser-scale features 
$F_{s-1,1}, F_{s-1,2}$ and $C_{s-1}$ 
for $s \in \{2,3,4\}$. Thereby, the consolidation is realized by stacking the upsampled coarser features and the intermediate finer features and aggregating them via a residual unit. 

This choice of architecture, however, is problematic for the computation of both image features and context features, due to the final ReLU activation applied in the residual unit. 
In the case of image features, this activation limits the range and therefore the expressiveness of the features of both images that are 
used in the computation of matching costs.
In the case of context features, using this consolidation module is also not advisable, since the initial hidden state is computed as a Tanh activated split of the multi-scale context features after the consolidation, which is already activated via ReLU and does not contain negative values that Tanh expects for computing the hidden-state of the GRU. 

We therefore improved the multi-scale feature consolidation module by
adding an activation-free convolution layer after the residual unit at each scale, a small but effective modification to regain expressiveness. 
At the same time, using a leaner residual unit for the consolidation in the context encoder allowed us to reduce the size of entire optical flow
network 
by 28\% without harming the performance (see suppl. material).
Our improved multi-scale feature consolidation is illustrated in the upper part of \cref{fig:consolidation-GC}. Further details 
can be found in the suppl. material.

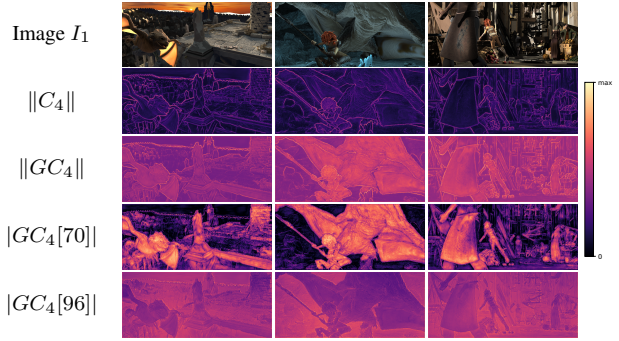
\begin{figure}[t]
    \centering
    \input{figures/feat_map_new}
    \caption{Visualization of context features. Top to bottom: reference frame, visualization of context $C_4$, $GC_4$, two channels of $GC_4$. Similar to \cite{xcit_2021_ali}, the $L_2$ norm of each feature map is shown as heat map. Each feature map is normalized individually.}
    \label{fig:feat_map}
\end{figure}
\subsection{Global Context Computation}
\label{subsec:global_context}
So far, we discussed how the multi-scale image and context features as well as the initial hidden-state per-scale are computed. Considering the context features $C_s$ 
shown in \cref{fig:consolidation-GC}, we now detail on how we compute our global context features. In this case, the goal is to obtain more attentive features that are
later on used for guiding the motion.
To this end, we perform 
an aggregation of the context features $C_s$ via its channel statistics using a self-attention XCiT layer \cite{xcit_2021_ali} (see lower part of \cref{fig:consolidation-GC}), which offers linear complexity w.r.t the number of tokens. This architectural choice allows a feasible aggregation of context over all the coarse-to-fine scales during the estimation.
Importantly, our adaption of XCiT differs from its original approach, where the XCiT layer is actually applied on a coarser representation of its input realized via explicit input patchification and then again upsampled to obtain the original resolution.
We, in contrast, apply the XCiT layer directly to the features in all coarse-to-fine scales, benefiting from scale-specific content.

To compute the global context, first, a sinusoidal positional embedding \cite{vaswani_2017_attention} is added to the context features $C_s$, yielding $C_{s_p}\in \mathbb{R}^{N\times d}$, where $N$ and $d$ denote the number of pixels at scale $s$ and number of feature channels, respectively. This step is then followed by a layer normalization $\mathcal{N}(C_{s_p})$ \cite{ba-layer-norm-2016} $\!$\footnote{For a more compact illustration we omitted both steps in \cref{fig:consolidation-GC}.}. 
At this point, to perform self-attention, all key ($K$), query ($Q$) and value features ($V$) features are computed from $\mathcal{N}(C_{s_p})$ via
\begin{equation}
\begin{aligned}
K_{C_s} &= \mathcal{N}(C_{s_p}) \cdot W_{s,k}, \\
Q_{C_s} &= \mathcal{N}(C_{s_p}) \cdot W_{s,q}, \\
V_{C_s} &= \mathcal{N}(C_{s_p}) \cdot W_{s,v},
\end{aligned}    
\end{equation}
where $W_{s,k}, W_{s,q}, W_{s,v} \in \mathbb{R}^{d \times d}$ are learned weight matrices.
\begin{figure}
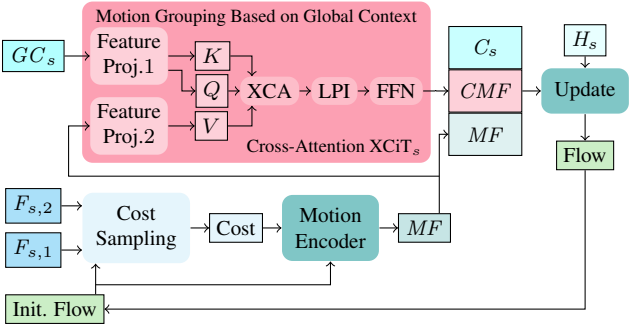

    \centering
    \include{figures/matching-block}
    \vspace{-9mm}
    \caption{Matching block of our CCMR approach for each scale. Motion features are guided based on global context iteratively and used in the flow update computation. The motion encoder and the update block are shared among scales.}
    \label{fig:CCMR}
\end{figure}
Before applying a cross-covariance attention step, 
 the channels of $K_{C_s}, Q_{C_s}, V_{C_s}\in \mathbb{R}^{N\times d}$ are reshaped into $h$ heads resulting in $K_{C_s}, Q_{C_s}, V_{C_s} \in \mathbb{R}^{N\times d/h \times h}$; cf.\ \cite{xcit_2021_ali}.
The cross-covariance attention then reads
\begin{equation}
\label{eq:attn}
    \text{XCA}(K_{C_s}, Q_{C_s}, V_{C_s}) = V_{C_s}\cdot \mathcal{S}(\hat{K}_{C_s}^{\top} \cdot \hat{Q}_{C_s}/\tau_s),
\end{equation}

\noindent where $\mathcal{S}$ denotes the Softmax function, $\tau_s$ is the learned temperature parameter and $\hat{K}_{C_s}$ and $\hat{Q}_{C_s}$ denote $L_2$-normalized key and query features, respectively. Note that the transpose operation and matrix multiplication in \cref{eq:attn} are along the first two dimensions.
Afterwards, a local patch interaction (LPI) layer and then a feed forward network (FFN) is applied. While the cross-covariance attention enables global interactions among channels in each head, LPI and FFN units provide explicit spatial interaction between tokens locally and communication across all channels \cite{xcit_2021_ali}, respectively.
The XCiT operations can be summarized as
\begin{alignat}{2}
\label{eq:weighted_sum1}
        C_{\text{int}_1} &{}={}& C_{s_p} &{}+{} \gamma_{1_s} \cdot \text{XCA($K_{C_s}, Q_{C_s}, V_{C_s}$)}, \\
        C_{\text{int}_2} &{}={}& C_{\text{int}_1} &{}+{} \gamma_{2_s} \cdot \text{LPI($\mathcal{N}(C_{\text{int}_1})$)}, \\
        \label{eq:weighted_sum2}
        GC_s &{}={}& C_{\text{int}_2} &{}+{} \gamma_{3_s} \cdot \text{FFN($\mathcal{N}(C_{\text{int}_2})$)},
\end{alignat}
where $C_{\text{int}_1}$ and $C_{\text{int}_2}$ are intermediate outcomes, $GC_s$ is the resulting global self-attentive context features, and $\gamma_{k_s}$ are learned scaling factors. The LPI module consists of two depth-wise convolutions and the FFN contains two linear layers. For simplicity and compactness, the weighted sum operations of \crefrange{eq:weighted_sum1}{eq:weighted_sum2} are not shown in \cref{fig:consolidation-GC}.

Before we discuss how the global self-attentive context features are used to guide motion features, let us first take a look at the impact of the global aggregation process. 
To this end, we visualize exemplary context maps in \cref{fig:feat_map}.
Row 2 and 3 show the channel-wise $L_2$ norm of the unaggregated context features $C_4$\footnote{For fairness, an independent model without global context computation for motion reasoning
was trained to generate the visualized $C_4$-maps.} 
and of our global context features $GC_4$. As one can see, in the case of $C_4$ the initial context features show high activations only mainly at edges, while in $GC_4$ the attention of the global context is spread over entire objects. Additionally, Row 4 and 5 depict the magnitude of two channels (70 and 96) of $GC_4$. While the former attends to the foreground, the latter has large activations in the background.

\subsection{Global-Context Based Motion Grouping}

\label{subsec:matching}

\begin{figure}
    \centering
    \includegraphics[width=0.9\columnwidth]{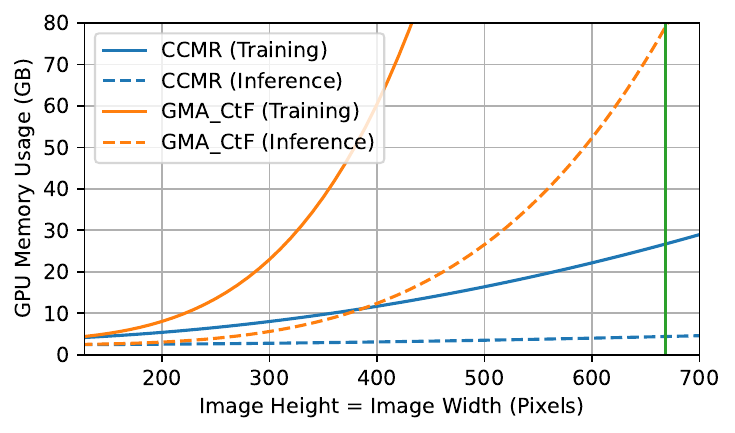}
    \vspace{-3mm}
    \caption{Comparison of memory consumption for our approach and the direct extension of GMA to multiple scales. X-axis shows height or width of squared input images.}
    \label{fig:memcheck}
\end{figure}

\input{figures/examples_new}

In this section, we finally elaborate on the flow estimation via global-context based motion grouping based on the image features $F_{s,1}, F_{s,2}$, the initial and global context features $C_s$ and $GC_s$ and the initial hidden state $H_s$ which where explained in \cref{subsec:feat_consolidation} and \cref{subsec:global_context}.
First, based on the initial flow in the first iteration (or the updated flow in the next iterations), matching costs in a neighborhood are computed from image features ($F_{s,1}, F_{s,2}$). Note that unlike RAFT\cite{teedRAFTRecurrentAllPairs2020}, the all-pairs cost volume is not pre-computed, due to the large memory consumption for finer scales. The computed costs together with the current flow estimate are then processed via a motion encoder which outputs motion features (see \cref{fig:CCMR}) that are eventually used in the GRU to compute the flow update.

 When computing the recurrent flow updates, additionally including global aggregated motion features based on context features 
can help resolve ambiguities in occluded areas.
This is intuitive, because the motion of occluded pixels from a partially non-occluded object can typically be inferred from the motion of its non-occluded pixels. 
In contrast to GMA \cite{jiangLearningEstimateHidden2021}, which realizes this additional step with the means of a single token cross-attention \cite{vaswani_2017_attention} using coarse non-aggregated context to aggregate motion features in a single-scale setting, we follow our efficient strategy based on global channel statistics from the computation of the global context, which is performed on all coarse-to-fine scales. 
More precisely, we perform the motion grouping by applying a cross-attention XCiT layer, on the global self-attentive context $GC_s$ and motion features $MF$. 
Thus, we compute key and query features from our global context features $GC_s$ and value features from the motion features directly from each scale without explicit patchification. 
After applying XCA, LPI and FFN to the key, query and value features, the context-guided motion features ($CMF$ in \cref{fig:CCMR}), the context features $C_s$ and the initial motion features $MF$ are stacked and passed through the recurrent unit to compute the flow update iteratively.

Note that applying token cross-attention\cite{vaswani_2017_attention} to perform motion aggregation in high-resolution coarse-to-fine schemes is memory-wise infeasible. 
\cref{fig:memcheck} shows that a simple GMA-based coarse-to-fine variant with the finest scale at $(\frac{h}{2}, \frac{w}{2})$ requires an enormous amount of VRAM. Even for comparably small images (400 $\times$ 400 pixels) it needs more VRAM for inference than CCMR requires for training. 
Note that inference for Sintel images with $436\times1024 \approx 668^2$ pixels in this case requires around 80 GB, while our approach needs less than 5 GB.
In the suppl. material, we provide additional details on this comparison.

\begin{table*}[t!]
    \caption{Comparison to the SOTA and the baselines on Sintel (test) and KITTI (test) using the AEPE (Sintel) and the Fl error (KITTI).\linebreak Mat = matched. Unmat = unmatched. Best result in bold, second best result is underlined. Results taken from the corresponding papers.}
	\label{eval:tab:literature}
	\centering
    \scalebox{0.85}{
    \begin{tabular}{l r c >{\columncolor{mygray}[6pt][4pt]}cccccc c >{\columncolor{mygray}[6pt][4pt]}cccccc l r c >{\columncolor{mygray}[6pt][4pt]}cccc c}
		\toprule
		 &\!\!\!\!&\;& \multicolumn{6}{c}{\!\!\!\!\!\!Sintel Clean (test)} 
		 &\!\!\!\!& \multicolumn{6}{c}{\!\!\!\!\!\!Sintel Final (test)} 
		 &&\;&\!\!\!\!& \multicolumn{3}{c}{KITTI (test)}
		  \\\cline{4-8}\cline{11-15}\cline{20-22}
        \\[-11pt]
		 \rowcolor[gray]{1.0}
   Method &&\!\!\!\!& {\;All$\downarrow$\;}
		 &\!\!\!\!\!\!\!\!\!\!\!& {\;Mat$\downarrow$\;}
		 &\!\!\!\!\!\!\!\!\!\!\!\!& \!\!\!\!\!\!\!\!\!\!{\;Unmat$\downarrow$\;}\!\!\!\!\!\!\!\!\!\!&
	  	 &\!\!\!\!& {\;All$\downarrow$\;}
		 &\!\!\!\!\!\!\!\!\!\!\!& {\;Mat$\downarrow$\;}
		 &\!\!\!\!\!\!\!\!\!\!\!\!& \!\!\!\!\!\!\!\!\!\!{\;Unmat$\downarrow$\;}\!\!\!\!\!\!\!\!\!\!&
		 & Method &\!\!\!\!&\!\!\!\!& {\;Fl-all$\downarrow$\;}
		 &\!\!\!\!\!\!\!\!\!\!\!& {\;Fl-noc$\downarrow$\;}
        \\
		\midrule
        
        RAFT~\cite{teedRAFTRecurrentAllPairs2020}  
        \tiny{\textcolor{gray}{ECCV'20}}
        \!\!\!\!\!\!\!\!\!\!\!\!\!\!\!\!\!\!\!\!\!\!\!\!\!\!\!\!\!\!\!\!\!\!\!\!\! &
        \!\!\!\!\!\!\!\!&        
		\!\!\!\!&
        1.61                          &\!\!\!\!\!\!\!\!\!&
		0.62                          &\!\!\!\!\!\!\!\!\!\!\!\!\!\!\!&
		\phantom{0}9.65               &&
        \!\!\!\!&
		2.86                          &\!\!\!\!\!\!\!\!\!&
		1.41                          &\!\!\!\!\!\!\!\!\!\!\!\!\!\!\!&
		14.68                         &&
        RAFT~\cite{teedRAFTRecurrentAllPairs2020} 
        \tiny{\textcolor{gray}{ECCV'20}}
        \!\!\!\!\!\!\!\!\!\!\!\!\!\!\!\!\!\!\!\!\!\!\!\!\!\!\!\!\!\!\!\!\!\!\!\!\! &
        \!\!\!\!&
		\!\!\!\!&
		5.10                          &\!\!\!\!\!\!\!\!\!\!\!&
		3.07
        \\
        RAFT-it~\cite{sunDisentanglingArchitectureTraining2022} 
        \tiny{\textcolor{gray}{ECCV'22}}
        \!\!\!\!\!\!\!\!\!\!\!\!\!\!\!\!\!\!\!\!\!\!\!\!\!\!\!\!\!\!\!\!\!\!\!\!\! &
        \!\!\!\!&        
		\!\!\!\!&
        1.55                       &\!\!\!\!\!\!\!\!\!\!\!&
		0.61                          &\!\!\!\!\!\!\!\!\!\!\!\!\!\!\!&
		\phantom{0}9.24               &&
        \!\!\!\!&
		2.90                          &\!\!\!\!\!\!\!\!\!\!\!&
		1.41                          &\!\!\!\!\!\!\!\!\!\!\!\!\!\!\!&
		15.03                         &&		
        GMA$\hspace{-0.3mm}^{\boldsymbol\dagger}\hspace{-1.05mm}$~\cite{jiangLearningEstimateHidden2021} 
        \tiny{\textcolor{gray}{ICCV'21}}
        \!\!\!\!\!\!\!\!\!\!\!\!\!\!\!\!\!\!\!\!\!\!\!\!\!\!\!\!\!\!\!\!\!\!\!\!\! &
        \!\!\!\!&
		\!\!\!\!&
		4.93                         &\!\!\!\!\!\!\!\!\!\!\!&
		2.90
        \\		
        DIP~\cite{zhengDIPDeepInverse2022}
        \tiny{\textcolor{gray}{CVPR'22}}
        \!\!\!\!\!\!\!\!\!\!\!\!\!\!\!\!\!\!\!\!\!\!\!\!\!\!\!\!\!\!\!\!\!\!\!\!\! &
		\!\!\!\!&
        \!\!\!\!&
        1.44                          &\!\!\!\!\!\!\!\!\!\!\!&
		0.52                          &\!\!\!\!\!\!\!\!\!\!\!\!\!\!\!&
		\phantom{0}8.92               &&
        \!\!\!\!&
		2.83                          &\!\!\!\!\!\!\!\!\!\!\!&
		1.28                          &\!\!\!\!\!\!\!\!\!\!\!\!\!\!\!&
		15.49                         &&
        MS-RAFT~\cite{jahediMultiScaleRAFTCombining2022}
        \tiny{\textcolor{gray}{ICIP'22}}
        \!\!\!\!\!\!\!\!\!\!\!\!\!\!\!\!\!\!\!\!\!\!\!\!\!\!\!\!\!\!\!\!\!\!\!\!\! &
        \!\!\!\!&
		\!\!\!\!&
		4.88                          &\!\!\!\!\!\!\!\!\!\!\!&
		2.80 
        \\
        RAFT-OCTC~\cite{jeongImposingConsistencyOptical2022} 
        \tiny{\textcolor{gray}{CVPR'22}}
        \!\!\!\!\!\!\!\!\!\!\!\!\!\!\!\!\!\!\!\!\!\!\!\!\!\!\!\!\!\!\!\!\!\!\!\!\! &
        \!\!\!\!&        
		\!\!\!\!&
        1.42                          &\!\!\!\!\!\!\!\!\!\!\!&
		0.54                          &\!\!\!\!\!\!\!\!\!\!\!\!\!\!\!&
		\phantom{0}8.57               &&
        \!\!\!\!&
		2.57                          &\!\!\!\!\!\!\!\!\!\!\!&
		1.24                          &\!\!\!\!\!\!\!\!\!\!\!\!\!\!\!&
		13.44                         &&	      
        FlowFormer~\cite{huangFlowFormer2022}
        \tiny{\textcolor{gray}{ECCV'22}}
        \!\!\!\!\!\!\!\!\!\!\!\!\!\!\!\!\!\!\!\!\!\!\!\!\!\!\!\!\!\!\!\!\!\!\!\!\! &
        \!\!\!\!&
		\!\!\!\!&
		4.68                          &\!\!\!\!\!\!\!\!\!\!\!&
		2.69
	    \\                                                          
        GMA
        ~\cite{jiangLearningEstimateHidden2021} 
        \tiny{\textcolor{gray}{ICCV'21}}
        \!\!\!\!\!\!\!\!\!\!\!\!\!\!\!\!\!\!\!\!\!\!\!\!\!\!\!\!\!\!\!\!\!\!\!\!\! &
        \!\!\!\!&        
		\!\!\!\!&
		1.39                          &\!\!\!\!\!\!\!\!\!\!\!&
		0.58                          &\!\!\!\!\!\!\!\!\!\!\!\!\!\!\!&
		\phantom{0}7.96               &&
        \!\!\!\!&  
		2.47                          &\!\!\!\!\!\!\!\!\!\!\!&
		1.24                          &\!\!\!\!\!\!\!\!\!\!\!\!\!\!\!&
		12.50                         &&
        MatchFlow\_GMA~\cite{dongMatchFlowGMA2023}
        \tiny{\textcolor{gray}{CVPR'23}}
        \!\!\!\!\!\!\!\!\!\!\!\!\!\!\!\!\!\!\!\!\!\!\!\!\!\!\!\!\! &
        \!\!\!\!&
		\!\!\!\!&
	    4.63                          &\!\!\!\!\!\!\!\!\!\!\!&
		2.77     
        \\        
        MS-RAFT~\cite{jahediMultiScaleRAFTCombining2022} 
        \tiny{\textcolor{gray}{ICIP'22}}
        \!\!\!\!\!\!\!\!\!\!\!\!\!\!\!\!\!\!\!\!\!\!\!\!\!\!\!\!\!\!\!\!\!\!\!\!\! &
        \!\!\!\!&        
		\!\!\!\!&
        1.37                          &\!\!\!\!\!\!\!\!\!\!\!&
		0.48                          &\!\!\!\!\!\!\!\!\!\!\!\!\!\!\!&
		\phantom{0}8.67               &&
        \!\!\!\!&
		2.67                          &\!\!\!\!\!\!\!\!\!\!\!&
		1.20                          &\!\!\!\!\!\!\!\!\!\!\!\!\!\!\!&
		14.70                         &&
        KPA-Flow~\cite{luoLearningOpticalFlow2022}
        \tiny{\textcolor{gray}{CVPR'22}}
        \!\!\!\!\!\!\!\!\!\!\!\!\!\!\!\!\!\!\!\!\!\!\!\!\!\!\!\!\!\!\!\!\!\!\!\!\! &
        \!\!\!\!&
		\!\!\!\!&
		4.60                          &\!\!\!\!\!\!\!\!\!\!\!&
		2.82   
		\\  
        KPA-Flow~\cite{luoLearningOpticalFlow2022}
        \tiny{\textcolor{gray}{CVPR'22}}
        \!\!\!\!\!\!\!\!\!\!\!\!\!\!\!\!\!\!\!\!\!\!\!\!\!\!\!\!\!\!\!\!\!\!\!\!\! &
        \!\!\!\!&        
		\!\!\!\!&
        1.35                          &\!\!\!\!\!\!\!\!\!\!\!&
		0.50                          &\!\!\!\!\!\!\!\!\!\!\!\!\!\!\!&
		\phantom{0}8.23               &&
        \!\!\!\!&
		2.27                          &\!\!\!\!\!\!\!\!\!\!\!&
		1.09                          &\!\!\!\!\!\!\!\!\!\!\!\!\!\!\!&
		11.89                         &&
        FlowFormer++~\cite{shiFlowFormerplusplus2023}
        \tiny{\textcolor{gray}{CVPR'23}}
        \!\!\!\!\!\!\!\!\!\!\!\!\!\!\!\!\!\!\!\!\!\!\!\!\!\!\!\!\!\!\!\!\!\!\!\!\! &
        \!\!\!\!&
		\!\!\!\!&
		4.52                          &\!\!\!\!\!\!\!\!\!\!\!&
		-     
        \\
       
        MS-RAFT+~\cite{jahediMSRAFTplus2022} 
        \tiny{\textcolor{gray}{arXiv'22}}
        \!\!\!\!\!\!\!\!\!\!\!\!\!\!\!\!\!\!\!\!\!\!\!\!\!\!\!\!\!\!\!\!\!\!\!\!\! &
        \!\!\!\!&        
		\!\!\!\!&
        1.23                          &\!\!\!\!\!\!\!\!\!\!\!&
		0.40                          &\!\!\!\!\!\!\!\!\!\!\!\!\!\!\!&
		\phantom{0}8.02               &&
        \!\!\!\!&
		2.68                          &\!\!\!\!\!\!\!\!\!\!\!&
		1.28                          &\!\!\!\!\!\!\!\!\!\!\!\!\!\!\!&
		14.12                         &&        
        GMFlow+~\cite{xuUnifyingFlowStereo2022}
        \tiny{\textcolor{gray}{TPAMI'23}}
        \!\!\!\!\!\!\!\!\!\!\!\!\!\!\!\!\!\!\!\!\!\!\!\!\!\!\!\!\!\!\!\!\!\!\!\!\! &
        \!\!\!\!&
		\!\!\!\!&
		4.49                          &\!\!\!\!\!\!\!\!\!\!\!&
		2.40           
        \\
        AnyFlow~\cite{jungAnyFlow2023}
        \tiny{\textcolor{gray}{CVPR'23}}
        \!\!\!\!\!\!\!\!\!\!\!\!\!\!\!\!\!\!\!\!\!\!\!\!\!\!\!\!\!\!\!\!\!\!\!\!\! &
        \!\!\!\!&        
		\!\!\!\!&
        1.23                          &\!\!\!\!\!\!\!\!\!\!\!&
		-                          &\!\!\!\!\!\!\!\!\!\!\!\!\!\!\!&
		-              &&
        \!\!\!\!&
		  2.44                         &\!\!\!\!\!\!\!\!\!\!\!&
		-                          &\!\!\!\!\!\!\!\!\!\!\!\!\!\!\!&
		-                         &&
        AnyFlow~\cite{jungAnyFlow2023}
        \tiny{\textcolor{gray}{CVPR'23}}
        \!\!\!\!\!\!\!\!\!\!\!\!\!\!\!\!\!\!\!\!\!\!\!\!\!\!\!\!\! &
        \!\!\!\!&
		\!\!\!\!&  
		4.41                          &\!\!\!\!\!\!\!\!\!\!\!&
		2.69                                            
        \\
       
        FlowFormer~\cite{huangFlowFormer2022}
        \tiny{\textcolor{gray}{ECCV'22}}
        \!\!\!\!\!\!\!\!\!\!\!\!\!\!\!\!\!\!\!\!\!\!\!\!\!\!\!\!\!\!\!\!\!\!\!\!\! &
        \!\!\!\!&		
		\!\!\!\!&
        1.14                          &\!\!\!\!\!\!\!\!\!\!\!&
		-                             &\!\!\!\!\!\!\!\!\!\!\!\!\!\!\!&
		-                             &&
        \!\!\!\!&
		  2.18                          &\!\!\!\!\!\!\!\!\!\!\!&
		-                             &\!\!\!\!\!\!\!\!\!\!\!\!\!\!\!&
		-                             &&             
        RAFT-OCTC~\cite{jeongImposingConsistencyOptical2022} 
        \tiny{\textcolor{gray}{CVPR'22}}
        \!\!\!\!\!\!\!\!\!\!\!\!\!\!\!\!\!\!\!\!\!\!\!\!\!\!\!\!\!\!\!\!\!\!\!\!\! &
        \!\!\!\!&
		\!\!\!\!&
		4.33                          &\!\!\!\!\!\!\!\!\!\!\!&
		-
        \\
       
        MatchFlow\_GMA~\cite{dongMatchFlowGMA2023}
        \tiny{\textcolor{gray}{CVPR'23}}
        \!\!\!\!\!\!\!\!\!\!\!\!\!\!\!\!\!\!\!\!\!\!\!\!\!\!\!\!\!\!\!\!\!\!\!\!\! &
        \!\!\!\!&        
		\!\!\!\!&
        1.16                          &\!\!\!\!\!\!\!\!\!\!\!&
		0.43                          &\!\!\!\!\!\!\!\!\!\!\!\!\!\!\!&
		\phantom{0}7.13               &&
        \!\!\!\!&
		  2.37                          &\!\!\!\!\!\!\!\!\!\!\!&
		1.06                          &\!\!\!\!\!\!\!\!\!\!\!\!\!\!\!&
		13.07                         &&
        RAFT-it~\cite{sunDisentanglingArchitectureTraining2022}
        \tiny{\textcolor{gray}{ECCV'22}}
        \!\!\!\!\!\!\!\!\!\!\!\!\!\!\!\!\!\!\!\!\!\!\!\!\!\!\!\!\!\!\!\!\!\!\!\!\! &
        \!\!\!\!&
		\!\!\!\!&
		4.31                          &\!\!\!\!\!\!\!\!\!\!\!&
		  -                       
        \\         
        FlowFormer++~\cite{shiFlowFormerplusplus2023}
        \tiny{\textcolor{gray}{CVPR'23}}
        \!\!\!\!\!\!\!\!\!\!\!\!\!\!\!\!\!\!\!\!\!\!\!\!\!\!\!\!\!\!\!\!\!\!\!\!\! &
        \!\!\!\!&
        \!\!\!\!&
        \underline{1.07}              &\!\!\!\!\!\!\!\!\!\!\!&
		0.39                          &\!\!\!\!\!\!\!\!\!\!\!\!\!\!\!&
		\phantom{0}{\bf 6.64}         &&
        \!\!\!\!&
		  {\bf 1.94}                    &\!\!\!\!\!\!\!\!\!\!\!&
		{\bf 0.88}                    &\!\!\!\!\!\!\!\!\!\!\!\!\!\!\!&
		{\bf 10.63}                   && 
        DIP~\cite{zhengDIPDeepInverse2022}
        \tiny{\textcolor{gray}{CVPR'22}}
        \!\!\!\!\!\!\!\!\!\!\!\!\!\!\!\!\!\!\!\!\!\!\!\!\!\!\!\!\!\!\!\!\!\!\!\!\! &
		\!\!\!\!&
        \!\!\!\!&
		4.21                          &\!\!\!\!\!\!\!\!\!\!\!&
		2.43                              
        \\ 
        GMFlow+~\cite{xuUnifyingFlowStereo2022}
        \tiny{\textcolor{gray}{TPAMI'23}}
        \!\!\!\!\!\!\!\!\!\!\!\!\!\!\!\!\!\!\!\!\!\!\!\!\!\!\!\!\!\!\!\!\!\!\!\!\! &
        \!\!\!\!&        
		\!\!\!\!&
        {\bf 1.03}                    &\!\!\!\!\!\!\!\!\!\!\!&
		\underline{0.34}              &\!\!\!\!\!\!\!\!\!\!\!\!\!\!\!&
		\phantom{0}\underline{6.68}   &&
        \!\!\!\!&
		2.37                          &\!\!\!\!\!\!\!\!\!\!\!&
		1.10                          &\!\!\!\!\!\!\!\!\!\!\!\!\!\!\!&
		12.74                         &&                        
        MS-RAFT+~\cite{jahediMSRAFTplus2022}
        \tiny{\textcolor{gray}{arXiv'23}}
        \!\!\!\!\!\!\!\!\!\!\!\!\!\!\!\!\!\!\!\!\!\!\!\!\!\!\!\!\!\!\!\!\!\!\!\!\! &
        \!\!\!\!&
		\!\!\!\!&
		4.15                          &\!\!\!\!\!\!\!\!\!\!\!&
		\underline{2.18}  
        \\ 
        
       	\midrule
        {\bf CCMR (ours)}
        \!\!\!\!\!\!\!\! &
        \!\!\!\!&
        \!\!\!\!&
		1.19                          &\!\!\!\!\!\!\!\!\!\!\!&
		0.41                          &\!\!\!\!\!\!\!\!\!\!\!\!\!\!\!&
		\phantom{0}7.50               &&
        \!\!\!\!&
		2.14                          &\!\!\!\!\!\!\!\!\!\!\!&
		1.00                          &\!\!\!\!\!\!\!\!\!\!\!\!\!\!\!&
		\underline{11.44}                         &&		
        {\bf CCMR (ours)}
        \!\!\!\!&
        \!\!\!\!&
		\!\!\!\!&
		\underline{4.04}              &\!\!\!\!\!\!\!\!\!\!\!&
		2.44
        \\
        \;\;\;\; vs. GMA
        \!\!\!\!\!\!\!\! &
        \!\!\!\!&        
		\!\!\!\!&
		\!+14.4\%\!                        &\!\!\!\!\!\!\!\!\!\!\!&
		\!+29.3\%\!                        &\!\!\!\!\!\!\!\!\!\!\!\!\!\!\!&
        \!\phantom{0}+5.8\%\!              &&
        \!\!\!\!&
		\!+13.3\%\!                        &\!\!\!\!\!\!\!\!\!\!\!&
		\!+19.4\%\!                        &\!\!\!\!\!\!\!\!\!\!\!\!\!\!\!&
        \!\phantom{0}+8.5\%\!              &&
        \;\;\;\; vs. GMA
        \!\!\!\!&
        \!\!\!\!&
		\!\!\!\!&
		\!+18.1\%\!                        &\!\!\!\!\!\!\!\!\!\!\!&
		\!+15.9\%\!
        \\
        \;\;\;\; vs. MS-RAFT
        \!\!\!\!\!\!\!\! &
        \!\!\!\!&        
		\!\!\!\!&
		\!+13.1\%\!                        &\!\!\!\!\!\!\!\!\!\!\!&
		\!+14.6\%\!                        &\!\!\!\!\!\!\!\!\!\!\!\!\!\!\!&
		\!+13.5\%\!                        &&
        \!\!\!\!&
		\!+19.9\%\!                        &\!\!\!\!\!\!\!\!\!\!\!&
		\!+16.7\%\!                        &\!\!\!\!\!\!\!\!\!\!\!\!\!\!\!&
		  \!+22.2\%\!                        &&
        \;\;\;\; vs. MS-RAFT
        \!\!\!\!&
        \!\!\!\!&
		\!\!\!\!&
		\!+17.2\%\!                        &\!\!\!\!\!\!\!\!\!\!\!&
		\!+12.9\%\!
        \\
        \midrule
        {\bf CCMR+ (ours)}
        \!\!\!\!\!\!\!\! &
        \!\!\!\!&        
		\!\!\!\!&
		\underline{1.07}                   &\!\!\!\!\!\!\!\!\!\!\!&
		{\bf 0.31}                         &\!\!\!\!\!\!\!\!\!\!\!\!\!\!\!&
		\phantom{0}7.24                    &&
        \!\!\!\!&
		\underline{2.10}                   &\!\!\!\!\!\!\!\!\!\!\!&
		\underline{0.93}                   &\!\!\!\!\!\!\!\!\!\!\!\!\!\!\!&
		11.58                &&
        {\bf CCMR+ (ours)}
        \!\!\!\! &
        \!\!\!\!&
		\!\!\!\!&
		{\bf 3.86}                         &\!\!\!\!\!\!\!\!\!\!\!&
		{\bf 2.07}
        \\
        \;\;\;\; vs. GMA
        \!\!\!\!\!\!\!\! &
        \!\!\!\!&        
		\!\!\!\!&
		\!+23.0\%\!                        &\!\!\!\!\!\!\!\!\!\!\!&
		\!+46.6\%\!                        &\!\!\!\!\!\!\!\!\!\!\!\!\!\!\!&
		\!\phantom{0}+9.1\%\!              &&
        \!\!\!\!&
		\!+15.0\%\!                        &\!\!\!\!\!\!\!\!\!\!\!&
		\!+25.0\%\!                        &\!\!\!\!\!\!\!\!\!\!\!\!\!\!\!&
		\!\phantom{0}+7.4\%\!              &&
        \;\;\;\; vs. GMA
        \!\!\!\! &
        \!\!\!\!&
		\!\!\!\!&
		\!+21.7\%\!                        &\!\!\!\!\!\!\!\!\!\!\!&
		\!+28.6\%\!
        \\
        \;\;\;\; vs. MS-RAFT+
        \!\!\!\!\!\!\!\! &
        \!\!\!\!&        
		\!\!\!\!&
		\!+13.0\%\!                        &\!\!\!\!\!\!\!\!\!\!\!&
		\!+22.5\%\!                        &\!\!\!\!\!\!\!\!\!\!\!\!\!\!\!&
		\!\phantom{0}+9.7\%\!              &&
        \!\!\!\!&
		\!+21.6\%\!                        &\!\!\!\!\!\!\!\!\!\!\!&
		\!+27.3\%\!                        &\!\!\!\!\!\!\!\!\!\!\!\!\!\!\!&
		  \!+18.0\%\!                        &&
        \;\;\;\; vs. MS-RAFT+
        \!\!\!\! &
        \!\!\!\!&
		\!\!\!\!&
		\!\phantom{0}+7.0\%\!              &\!\!\!\!\!\!\!\!\!\!\!&
		\!\phantom{0}+5.1\%\!
        \\
		\bottomrule\\[-2mm]         
     \multicolumn{22}{c}{$^{\boldsymbol\dagger}\hspace{-1.05mm}$ {
     GMA uses different architectures on Sintel and KITTI. 
    Using the Sintel architecture for KITTI yields Fl-all of 5.15.
     }
     } 
     
	\end{tabular}
 }
 \vspace{0mm}
\end{table*}

\begin{table}
\begin{center}
{\caption{Pre-training results. 
Results on {\em Sintel (train)} and {\em KITTI (train)} after pre-training with {\em Chairs} and {\em Things}. '-': not available.
}
\label{tab:pre-train}}
\vspace{-2mm}
\footnotesize
\begin{tabular}{l cc c}
\hline
\toprule
  & \multicolumn{2}{c}{Sintel (train)} & \multicolumn{1}{c}{KITTI (train)} \\
\cmidrule(r){2-3} \cmidrule(l){4-4}
 Method & Clean$\downarrow$ & Final$\downarrow$ & Fl$\downarrow$\\
 \midrule
  RAFT & 1.43 & 2.71 & 17.4\\
  GMA & 1.30 & 2.74 &  17.1\\
  DIP & 1.30 & 2.82 &  13.73\\
  KPA-flow &1.28 &2.68 & 15.9\\
  AnyFlow & 1.17 & 2.58 & \underline{13.01} \\
  MS-RAFT & 1.13 & 2.60 & -\\
  MatchFlow\_GMA & 1.03 & 2.45& 15.6\\
  Flowformer++ & \textbf{0.90} & \textbf{2.30} & 14.13\\
  GMFlow+ & \underline{0.91} &  - & -\\
  \midrule
  CCMR (ours)  & 1.07 & 2.40 & 13.30\\
  CCMR+ (ours)  & 0.98 & \underline{2.36}  & \textbf{12.84}\\
\bottomrule
\vspace{-2mm}
\end{tabular}

\end{center}
\vspace{-4mm}
\end{table}

\section{Experiments}
\label{experiments}

\subsection{Training Setup}
\label{subsec:experiments:training-setup}
We implemented our CCMR method in PyTorch.
Following MS-RAFT+, during training we used 4, 5, 5 and 6 GRU iterations from the coarsest to the finest scale, respectively.  
We trained the model for 120K iterations on Chairs \cite{Dosovitskiy2015_FlowNet}, 120K iterations on Things \cite{Mayer2016_FTH}, and 120K iterations on a mixed set containing Sintel \cite{butlerNaturalisticOpenSource2012}, KITTI 2015 \cite{Menze2015_KITTI}, HD1K \cite{Kondermann2016_HD1K}, and Things for {Sintel} evaluation. We then fine-tuned our model for 50K iterations on an automotive mixed set consisting of KITTI and Viper \cite{richterPlayingBenchmarks2017} for evaluating on {\em KITTI}.
We used a batch size of 9 in the first training phase, 6 for the remaining three phases. 
For supervision, we used the same multi-scale multi-iteration loss as \cite{jahediMultiScaleRAFTCombining2022} with increasing weights over scales and iterations. In this context, we used the $L_2$ loss for the first three training phases and the robust sample-wise loss \cite{jahediMultiScaleRAFTCombining2022} for the fine-tuning in the last phase.

Note that, besides the 4-scale model with finest matching scale at: ($\frac{h}{2},\frac{w}{2}$) based on to MS-RAFT+ \cite{jahediMSRAFTplus2022}, we also investigated a 3-scale version based on MS-RAFT\cite{jahediMultiScaleRAFTCombining2022} with finest scale at ($\frac{h}{4},\frac{w}{4}$). 
Consequently, in this section we rename the 4-scale  CCMR method explained so far to CCMR+, while referring to the smaller 3-scale version as CCMR.
Our CCMR and CCMR+ model with 10.8M and  11.5M parameters are $20\%$ and $28\%$ smaller than their MS-RAFT and MS-RAFT+ backbones with 13.5M and 16M parameters, respectively. 
This reduction of model size is due to our smaller modified context consolidation unit, which yielded better generalization results on average.
For detailed inference settings, memory requirements and runtime as well as for an ablation on the larger consolidation unit; see suppl. material.

\subsection{Results}
\label{subsec:experiments:results}
Let us now compare our CCMR(+) method to the state of the art and the underlying baselines (architectural baseline MS-RAFT(+) and conceptual baseline GMA), 
both in terms of accuracy and visual appearance.
To this end, we compare the optical flow accuracy on the {\em Sintel} \cite{butlerNaturalisticOpenSource2012} and {\em KITTI} \cite{Menze2015_KITTI} benchmarks\footnote{One unified architecture for each variant used on both benchmarks.} and additionally show pre-training generalization results. 
Finally, 
we also delineate the architectural impact from the impact of our fine-tuning for KITTI to outline 
the advantage of our architecture compared to the baselines.

\medskip
\noindent {\bf Comparison to the Baselines.}
Our results in \cref{eval:tab:literature} show large gains in 
both matched and unmatched regions, up to 46.6\% and 22.2\%, respectively. With large improvements against GMA in matched regions and against MS-RAFT(+) in unmatched regions, it becomes clear that our coarse-to-fine context-guided motion grouping is worthwhile. Specifically, our method outperforms GMA and MS-RAFT(+) by up to 23.0\% and 21.6\%, respectively. 
Also, pre-training generalization results show consistent improvements of our method over the baselines (see suppl. material).

\medskip
\noindent {\bf Comparison to the State of the Art.} Comparing the performance of our CCMR(+) models to the best reported results in the published literature,
\cref{eval:tab:literature} shows that our approach performs favorably.
While our 3-scale CCMR model achieves very good results on both benchmarks,
Our 4-scale CCMR+ models set a new state of the art on KITTI (3.86) and ranks second on Sintel Clean (1.07) and Final (2.10). In this context, our method shows good results in unmatched regions while it achieves an unprecedented accuracy for Sintel Clean (0.31) and KITTI (2.07) in matched areas---even outperforming the leading {\em scene flow} method on KITTI
(2.07 vs.\ 2.18, see \cite{liu2022camliflow} for results of CamLiRAFT).
Furthermore,
it is noteworthy that, in contrast to other methods from the literature our approach achieves excellent results on {\em both} Sintel {\em and} KITTI. For instance, GMFlow+ and FlowFormer++ achieve excellent results on Sintel clean (FlowFormer++ also on final), but are not among the SOTA on KITTI. RAFT-it and DIP in turn, that perform great 
on KITTI, offer less competitive results on Sintel.
Finally, we report our pre-training generalization results on Sintel (train) and KITTI (train) in \cref{tab:pre-train}. Also in this case, our models yield very good results. 

\medskip
\noindent {\bf Visual Comparison.}
The previous findings are also confirmed by a visual comparison of the results in \cref{fig:comparison}. Apart from our CCMR+ model, we also consider two recent SOTA methods 
for each benchmark 
as well as the GMA 
and MS-RAFT+ baseline, whenever available.
The corresponding flow fields show that our model preserves more details both in the foreground and the background (cf.\ boxes).

\medskip
\noindent {\bf Architectural Impact.}
As outlined in \cref{subsec:experiments:training-setup}, our models were fined-tuned for {KITTI}. While this is common practice in the literature (see, e.g., \cite{teedRAFTRecurrentAllPairs2020,sunDisentanglingArchitectureTraining2022,huangFlowFormer2022,Sun2022_NeurIPS,jeongImposingConsistencyOptical2022}), our multi-scale baselines MS-RAFT and MS-RAFT+ do not consider this additional fine-tuning step. Hence, to delineate the actual architectural improvements from the impact of the fine-tuning, we fined-tuned {both} our models (CCMR, CCMR+) and the multi-scale baselines (MS-RAFT, MS-RAFT+) on KITTI using the same training scheme for a fair comparison. 
To this end, we excluded a part of the KITTI training data in the preceding mixed training and the subsequent fine-tuning and used this missing data afterwards as {\em validation split} (see suppl. material).
The corresponding results in \cref{kitti_validation} show clear improvements from CCMR over MS-RAFT and CCMR+ over MS-RAFT+ as well as improvements when going from three to four scales. For completeness, we also included the result of our conceptual baseline, GMA, with the same training scheme. 

\begin{table}
\vspace{0mm}
\begin{center}
{\caption{Comparison of our CCMR(+) models with MS-RAFT(+) and GMA baselines on a {\em KITTI validation split} after {\em fine-tuning}. The same training protocol has been used in all cases such that observed differences are only due to the improved architecture.}\label{kitti_validation}}
\resizebox{\columnwidth}{!}{%
\begin{tabular}{lccccc}
\hline
\toprule
& GMA & MS-RAFT &CCMR &MS-RAFT+ & CCMR+\\
\midrule
KITTI-\textbf{val} & 5.01 & 5.21 &  4.75 & \underline{4.71} & \textbf{4.57} \\
\midrule
\#Scales&1&3&3&4&4\\ \midrule
Finest Res. & $\frac{1}{8}(h,w)$& $\frac{1}{4}(h,w)$ & $\frac{1}{4}(h,w)$ & $\frac{1}{2}(h,w)$ & $\frac{1}{2}(h,w)$ \\
\bottomrule
\end{tabular}
}
\end{center}
\vspace{-4mm}
\end{table}

\subsection{Ablations}
\label{sec:ablations}
After we have evaluated our models, we now ablate their underlying architecture. To this end, we 
compare the results of our method with different architectural configurations on {\em Sintel (train)}~\cite{butlerNaturalisticOpenSource2012} after pre-training the variants with {\em Chairs} \cite{Dosovitskiy2015_FlowNet} and {\em Things} \cite{Mayer2016_FTH}.
Due to the large computational effort, we performed the ablations on the 3-scale CCMR model. Finally, we also present results for our 4-scale CCMR+ model for the selected setting.

\medskip 
\noindent {\bf Architectural Components.}
In our first ablation, we study the impact of the improved feature consolidation unit and of the global-context based motion reasoning.
 The results in \cref{tab:IE_CMA} show that both concepts improve the estimation. While the context-based motion grouping has more impact on unmatched regions, the improved encoders have more influence in matched areas. 

\medskip 
\noindent {\bf Motion Grouping Based on Global Context.}
In the next ablation, we disentangle the impact of using global context features and motion grouping for components with
weights shared across scales or individual weights per scale; 
see \cref{tab:shared}. For the first two rows, no motion grouping was performed, and the global context was used directly in the recurrent unit for flow updates. In row three and four, no global context was computed and the motion grouping was done based on the un-aggregated context features.
In the remaining rows, we investigate the use of both concepts using shared or individual components across the coarse-to-fine scales, as well as going from the
3-scale CCMR model to the 4-scale CCMR+ model. 
The results show that both global context and motion grouping contribute to the improvements, with the best overall performance obtained by the model with individual units for each scale. Moreover, they demonstrate that going from three to four scales 
once again improves the results.

\begin{table}
\vspace{0mm}
\begin{center}
{\caption{Ablation on the {\em global-context based motion grouping} (GC-MG) and the {\em improved feature consolidation unit} (IFCU) in all, matched and unmatched regions.
Results on {\em Sintel (train)} after pre-training with {\em Chairs} and {\em Things}. Gray: CCMR configuration.
}\label{tab:IE_CMA}}
\resizebox{\columnwidth}{!}{%
\begin{tabular}{cccccccccc}
\hline
\toprule
  & & & \multicolumn{3}{c}{Sintel Clean (train)} & & \multicolumn{3}{c}{Sintel Final (train)} \\
\cline{4-6} \cline{8-10}
  GC-MG &  IFCU & & All$\downarrow$ & Mat$\downarrow$ & \!\!\!\!Unmat$\downarrow$\!\!\!\! & & All$\downarrow$ & Mat$\downarrow$ & \!\!\!\!Unmat$\downarrow$\!\!\!\!\\
\midrule
- & -  & & 1.18 & 0.47 & 10.20 & & 2.58 & 1.46 & 16.81\\
\checkmark & -  & & 1.10 & \underline{0.46} & \phantom{0}\underline{9.23} & & \underline{2.47} & \underline{1.42} & \underline{15.89}\\
\rowcolor{mygray} \checkmark & \checkmark  & &\textbf{1.07}& \textbf{0.43} & \phantom{0}\textbf{9.12}& &\textbf{2.40}& \textbf{1.35} &  \textbf{15.76}\\
\bottomrule
\end{tabular}
}
\end{center}
\vspace{-4mm}
\end{table}

\begin{table}
\begin{center}
{\caption{Ablation on {\em individual}$\;\!$/$\;\!${\em shared} modules for global context computation and context-based motion grouping across scales. 
Results on {\em Sintel (train)} after pre-training with {\em Chairs} and {\em Things.} 
Gray: CCMR and CCMR+ configuration.
}\label{tab:shared}}
\resizebox{\columnwidth}{!}{%
\begin{tabular}{ccccccc}
\hline
\toprule
 Global Context & Motion && \!\!\!\!\!Sintel (train)\!\!\!\! &\!\!& \!\!\!\!Sintel (train)\!\!\!\!\\
\cline{4-4} \cline{6-6}
  Computation & Grouping & \#Scales & Clean All$\downarrow$ &\!\!& Final All$\downarrow$\\
\midrule
shared & - &3& 1.11 &\!\!& 2.50 \\
individual & - &3& \underline{1.06} &\!\!& 2.53\\
- & shared &3&1.08&\!\!& 2.50\\
- & individual &3& 1.06 &\!\!& 2.48\\
\midrule
shared & shared &3&1.08&\!\!&\underline{2.45}\\
shared & individual &3& \textbf{1.03} &\!\!& \underline{2.45}\\
individual & shared &3&1.12&\!\!& 2.60\\
\rowcolor{mygray} individual & individual &3&1.07&\!\!& \textbf{2.40}\\
\midrule\midrule
individual & individual &3&\underline{1.07}&\!\!& \underline{2.40}\\
\rowcolor{mygray} individual & individual &4&\textbf{0.98}&\!\!& \textbf{2.36}\\
\bottomrule
\end{tabular}
}
\end{center}
\vspace{-5mm}
\end{table}
\section{Conclusion}

We presented CCMR, a high-accuracy optical flow approach that benefits from attention-based motion grouping concepts as well as from strong matching capabilities of high-resolution multi-scale neural networks. It relies on computing global context, which serves as guidance to perform motion reasoning across all 
scales via global channel statistics, and exploits an improved feature consolidation unit, that extracts more expressive features for matching.
Experiments show strong improvements in both occluded and non-occluded regions, resulting from the global-context based motion reasoning and the improved feature consolidation, respectively. In this way, our approach produces highly detailed flow fields in the foreground and background, while achieving state-of-the-art results on KITTI 2015 (Rank 1) and MPI Sintel (Rank 2 on Clean and Final).

\medskip 
\noindent {\bf Acknowledgements.}
The authors thank the Deutsche Forschungsgemeinschaft (DFG, German Research Foundation) -- Project-ID 251654672 -- TRR 161 (B04) and the International Max Planck Research School for Intelligent Systems (IMPRS-IS) for supporting Azin Jahedi.

{
\bibliographystyle{ieee_fullname}
\bibliography{egbib_short}
}

\cleardoublepage

\appendix

\section{Supplementary Material}
In the following, we provide additional details on our approach. 
Regarding the architecture, we further detail on the improved feature consolidation unit and how it is embedded in the whole feature extractor. We then show additional pre-training generalization results, comparing our approach to its baselines. After that we give additional details on the memory advantages of our approach compared to a simple coarse-to-fine GMA variant. 
 We then describe the validation set on KITTI that we used for delineate the architectural from the fine-tuning benefits.
Afterwards, we provide the inference settings of our approach and finally comment on practical aspects such as memory requirements and runtime.

\section{Improved Feature Extraction}
In \cref{subsec:feat_consolidation}, we gave a brief description of the U-Net style feature extractor\footnote{which is based on MS-RAFT(+)\cite{jahediMultiScaleRAFTCombining2022,jahediMSRAFTplus2022}}, and how the feature consolidation unit has been modified to increase its expressiveness and decrease its size. Let us now elaborate on the whole multi-scale feature extractor with the improved feature consolidation unit in more detail.

\cref{fig:unet} shows the architecture of the modified encoder for computing the image features. 
First, intermediate features are computed in a top-down manner which are denoted by $\hat{F}_2, \hat{F}_3$ and $\hat{F}_4$. Then to obtain multi-scale features,
the better-structured finer features
are enhanced by 
consolidating them with the deeper coarser-scale features.
To this end, bilinearly upsampled coarser features $F_{s-1}$ and intermediate finer features $\hat{F}_s$ are stacked and passed through a Res-Conv unit for consolidation. As the discussed in the paper, the Conv layer after the residual unit is activation-free. Note that two different encoders with the shown architecture are used for computing image and context features. While the one for the image features is shared among the two input images, the one for the context features has independent weights. \cref{fig:consolidation-GC} shows the feature consolidation unit  for each of the features $F_{s,1}, F_{s,2}, C_s$ but only for one scale (scale $s$), unlike what is illustrated in \cref{fig:unet} which shows the feature computation for different scales.
Note that the each of the dashed boxes in \cref{fig:unet} corresponds to feature consolidation in \cref{fig:consolidation-GC} at a fixed scale $s$.
\medskip

\begin{figure}[t!]
    \centering
    \scalebox{0.95}
    {
    \input{figures/encoder}
    }
    \vspace{-2mm}
    \caption{Our modified U-Net style feature extractor. Note that all convs shown in the figure are activation free. The red highlighted modules is our small but effective modification.}
    \vspace{-5mm}
    \label{fig:unet}
\end{figure}

In the multi-scale estimation framework,
to share the recurrent unit among all coarse-to-fine scales, the context features must have a fixed number of channels across scales (as they serve as input).
In \cite{jahediMultiScaleRAFTCombining2022,jahediMSRAFTplus2022}, this was realized by a large residual unit for consolidation to keep the output channels 256 at all scales. Using a leaner residual unit and converting the number of channels to 256 using the subsequent conv in the consolidation unit, our optical flow model became $28\%$ smaller. Our leaner residual unit has two residual layers with less intermediate and output channels. 
Following our ablation strategy from \cref{sec:ablations} which considers the 3-scale variant, our model with smaller residual unit yields pre-training generalization results of 1.07 for Sintel Clean and 2.40 for Final, which is a bit worse than 1.05, obtained by the larger model for Clean and better than its Final results: 2.49, respectively. Therefore the results of the smaller model were favorable on average.

\section{Pre-Training Generalization Results}
In \cref{subsec:experiments:results}, we compared the pre-training generalization results of our method to those of the baselines and of recent methods; see \cref{tab:pre-train}. The results reported in the respective publications, based on those methods using their {\em own} training schedules, are listed again in the upper part of \cref{tab:pre-train-supp}. Additionally, in the lower part of \cref{tab:pre-train-supp}, we now compare our results to the MS-RAFT(+) and GMA baselines for which, this time,  we applied the training schedule of {\em our} method, i.e., all methods have exactly the same training scheme. Also in this case, our CCMR(+) method outperforms the corresponding baselines, indicating that its improved pre-training generalization performance is not due to the different training setting.

\begin{table}[t!]
\begin{center}
{\caption{Pre-training results. 
Results on {\em Sintel (train)} and {\em KITTI (train)} after pre-training with {\em Chairs} and {\em Things}. 
}
\label{tab:pre-train-supp}}
\footnotesize
\begin{tabular}{l cc c}
\hline
\toprule
  & \multicolumn{2}{c}{Sintel (train)} & \multicolumn{1}{c}{KITTI (train)} \\
\cmidrule(r){2-3} \cmidrule(l){4-4}
 Method & Clean$\downarrow$ & Final$\downarrow$ & Fl$\downarrow$\\
 \midrule
  RAFT & 1.43 & 2.71 & 17.4\\
  GMA & 1.30 & 2.74 &  17.1\\
  DIP & 1.30 & 2.82 &  13.73\\
  KPA-flow &1.28 &2.68 & 15.9\\
  AnyFlow & 1.17 & 2.58 & \underline{13.01} \\
  MS-RAFT & 1.13 & 2.60 & -\\
  MatchFlow\_GMA & 1.03 & 2.45& 15.6\\
  Flowformer++ & \textbf{0.90} & \textbf{2.30} & 14.13\\
  GMFlow+ & \underline{0.91} &  - & -\\
  \midrule
  {\em GMA} & 1.36 & 2.74 & 17.51\\
  {\em MS-RAFT} & 1.18 & 2.58 & 13.84\\
  {\em MS-RAFT+} & 1.03 & 2.52  & 13.83\\
  {\em CCMR (ours)}  & 1.07 & 2.40 & 13.30\\
  {\em CCMR+ (ours)}  & 0.98 & \underline{2.36}  & \textbf{12.84}\\
\bottomrule
\end{tabular}
\end{center}
\end{table}

\section{CCMR vs.\ Coarse-to-Fine GMA}
\begin{figure}[t!]
    \centering
    \includegraphics[width=\columnwidth]{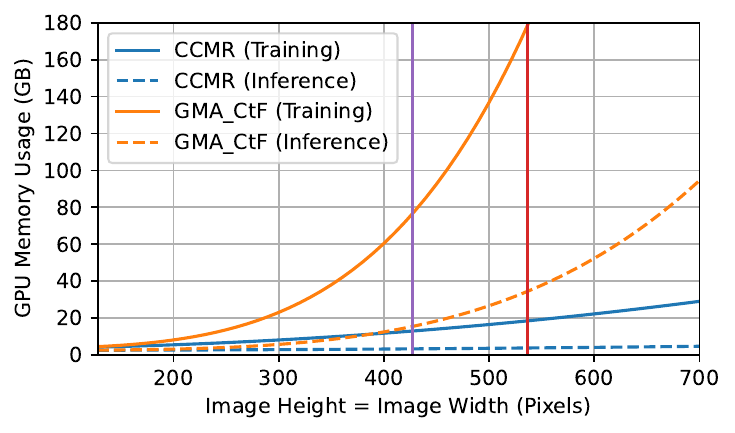}
    \caption{Comparison of memory consumption for our approach and the direct extension of GMA to multiple scales\protect\footnotemark. X-axis shows height or width of squared input images.}
    \label{fig:memcheck_sup}    
\end{figure}

In \cref{subsec:matching}, we outlined the infeasibility of a simple coarse-to-fine version of GMA. We demonstrated that even inference in such a network needs an enormous amounts of VRAM. In the following we add upon this by elaborating on the memory requirements to train such a network.
Considering patch sizes used by  RAFT's \cite{teedRAFTRecurrentAllPairs2020} training setting, the minimum and maximum size are $368 \times 496 \approx 427^2$ and $400 \times 720 \approx 536^2$, used in the first and second training stages, respectively. These patch sizes were also used for training GMA\cite{jiangLearningEstimateHidden2021} and our CCMR approach. Now let us take a look at the memory consumption for training \textbf{a single sample (batch of one)} shown in \cref{fig:memcheck_sup}.
 Considering the intersections of the solid orange line with the purple (patch size $\approx 427^2$) and red line (patch size of $\approx536^2$) shows that the coarse-to-fine version of GMA requires from 75 to about 180 GBs of VRAM. This means: training this network using GMA's training setting\footnote{GMA uses a batch of 8 samples for the first training phase and a batch size of 6 for the other training phases.} requires $8 \!\times\! 75 = 600$ GBs and about $6\!\times\!180=1080$ GBs for training the first stage on Chairs and second stage on Things, respectively, while our approach requires more than an order of magnitude less VRAM for training the second phase using the same patch size and batch size. In fact our approach needs at most 94 GBs which we distribute over three Nvidia A100 GPUs, each having 40 GBs of VRAM.
Note that in the measurements shown in \cref{fig:memcheck_sup}, the matching costs were computed on demand, without the computation of the all-pairs cost volume, therefore the consumed memory is not due to computing or keeping the 4D all-pairs cost volume in the VRAM, but for different partially attention-based coarse-to-fine calculations. 

Of course, when the memory requirements become critical, using smaller patches for training can be taken into account. However, since using larger patches for training provides more content for the network, reducing the patch size drastically during training is not advisable\cite{Bar-HaimW20_scopeflow}.

\footnotetext{Note that, for GMA\_CtF, we ran samples for inference and training for patch sizes which required memory up to the memory limit of our GPU---about 40 GBs ($\approx 352^2$ for training and $544^2$ for inference).
In this case, the memory consumption for larger patches was extrapolated using a quadratic function.}

\section{Validation Set for KITTI 2015}
\label{sec:kitti_val}
In \cref{subsec:experiments:results}, we delineated the architectural impact from the fine-tuning impact on KITTI.
To this end we split KITTI 2015 (train) into a training and a validation set. The validation set contained the following 50 sequences: \#0-\#29, \#60-\#69 and \#110-\#119 and the train-split consists of the remaining 150 samples. 
Using a fine-tuned version on such a split did not only allow us to compare the different baseline in a fair manner, but also to find effective number of iterations during inference.

\section{Inference Settings}
Similar to RAFT\cite{teedRAFTRecurrentAllPairs2020} and and many of its successors, such as GMA\cite{jiangLearningEstimateHidden2021}, we used different \textbf{inference} settings for the different benchmarks, however, as mentioned in the paper, we used one unified architecture\footnote{which was suggested by our ablations: last two rows in \cref{tab:shared}.} for each 3-scale and 4-scale variant for different pre-training and fine-tuning stages.

To compute the flow on the Sintel benchmark, we applied a cold warm-strategy as in MS-RAFT+\cite{jahediMSRAFTplus2022}, using 8 GRU iterations at the coarsest scale and 10 at other scales. For KITTI, 35 iterations at the two coarsest scales, 5 and 15 iterations at the next finer scales are performed, respectively, for the 4-scale model.
These iteration numbers were found empirically on the training set of Sintel for models pre-trained on Chairs and Things, and for KITTI on the KITTI validation split after fine-tuning the model on the training-split previously described in \cref{sec:kitti_val}.
In the case of our 3-scale model, we used 10, 15 and 20 GRU iterations for Sintel and 6, 18 and 30 iterations for KITTI from the coarsest to the finest scale, respectively.

Importantly, the optimal number of iterations for the best performance highly depends on the model.
Therefore, we investigated different combinations of GRU iterations per scale for both our method and our baselines and reported the best results 
in the lower part of \cref{tab:pre-train-supp}. In the same way, we performed our comparison to the baselines for the fine-tuned models on the KITTI validation split in \cref{kitti_validation}.
Therefore our clear advantage over the baselines is independent of the training or inference settings.
In our investigation, we found that in the case of KITTI, our model benefits from more GRU iterations than MS-RAFT(+), where MS-RAFT+ results degrades in accuracy using more than 10 GRU iterations per-scale while the accuracy of our method kept increasing. In the case of GMA, the accuracy barely changed by increasing the GRU iterations.

\section{Memory Requirements and Runtime}
Estimating the flow using our 4-scale model on Sintel-size images ($436 \times 1024$) takes about 0.9 seconds using 8, 10, 10 and 10 GRU iterations from the coarsest to the finest scale, respectively. This estimation requires 4.5 GBs of VRAM. Note that, in this setting, the matching costs are computed on-demand for each iteration.
Analogously, estimating the flow using our 3-scale model takes 0.55 seconds and requires 3.1 GBs of VRAM when performing 10, 15 and 20 GRU iterations from the coarsest to finest scale, 
respectively. Note that, in this case, as the finest matching scale of the 3-scale model is at $\frac{1}{4}(h, w)$ instead of at $\frac{1}{2}(h, w)$ in the 4-scale model, the computation of the all-pair cost volume is rather affordable. When doing so, the runtime reduces to 0.43 seconds, while the amount of required VRAM increases to 12.5 GB. 

\end{document}

%% file: macros.tex
\usepackage{graphicx}
\usepackage{latexsym}

\usepackage{booktabs}
\usepackage{colortbl}
\usepackage{mathtools}
\usepackage{tikz}
\usepackage[T1]{fontenc}
\usepackage{calc}
\usepackage{multirow}
\usepackage{siunitx}
\usepackage{subcaption}
\usepackage{nicefrac}
\usepackage{amssymb}
\newcommand{\skipping}[1]{}
\usepackage{pifont}
\usepackage{xcolor}
\usepackage{colortbl}
\usepackage{amsmath,amsfonts,amssymb}
\usepackage{cite}                   

\definecolor{mygray}{gray}{0.9}
\definecolor{celadon}{rgb}{0.67, 0.88, 0.69}
\usetikzlibrary{calc,backgrounds,decorations.pathreplacing,fit,positioning,shapes.geometric}
 \definecolor{cyan-context}{rgb}{0.0, 1.0, 1.0}
\colorlet{context color}{cyan-context!30}
\definecolor{scale color}{RGB}{0, 65, 145}
\colorlet{feature color}{cyan!30}
\definecolor{flow-color}{rgb}{0.66, 0.89, 0.63}
\colorlet{flow color}{flow-color!60}
\definecolor{image-color}{rgb}{1.0, 0.88, 0.21}
\colorlet{image color}{image-color!30}
\definecolor{blush}{rgb}{0.87, 0.36, 0.51}
\definecolor{KQV}{rgb}{1.0, 0.22, 0.0}
\definecolor{contrib-color}{rgb}{1.0, 0.11, 0.0}
\definecolor{darkpink}{rgb}{0.91, 0.33, 0.5}
\definecolor{blush}{rgb}{0.87, 0.36, 0.51}
\definecolor{brightmaroon}{rgb}{0.76, 0.13, 0.28}
\definecolor{contrib-color}{rgb}{0.98, 0.38, 0.5}
\definecolor{resconv}{rgb}{1.0, 0.72, 0.77}
\colorlet{cont_act}{gray!30}
\definecolor{bubblegum}{rgb}{0.99, 0.76, 0.8}
\definecolor{tealblue}{rgb}{0.21, 0.46, 0.53}
\definecolor{darkcyan}{rgb}{0.0, 0.55, 0.55}
\colorlet{shared color}{darkcyan!50}
\definecolor{almond}{rgb}{1.0, 0.92, 0.8}
\colorlet{res color}{almond!50}
\colorlet{resconv color}{resconv!50}
\colorlet{conv color}{res color}
\definecolor{int-feat-color}{rgb}{0.67, 0.9, 0.93}
\colorlet{intermediate context color}{context color!50}
\colorlet{intermediate feature color}{int-feat-color!50}
\colorlet{conv-no-act color}{contrib-color!50}
\colorlet{feat proj color}{contrib-color!30}
\colorlet{xcit color}{contrib-color!60}
\colorlet{kqv color}{contrib-color!30}
\tikzset{
    tensor/.style={draw,rectangle},
    operation/.style={rectangle,rounded corners},
    flow/.style={tensor, fill=flow color},
    context/.style={tensor, fill=context color, minimum size=1em},
    intermediate context/.style={context, fill=intermediate context color},
    context activation/.style={operation, fill=cont_act},
    feature/.style={tensor, fill=feature color, minimum size=1em},
    cost sampling/.style={operation, align=center, inner sep=5pt, fill=feature color!30},
    motion encoder/.style={operation, align=center, inner sep=5pt, fill=shared color},
    initial hidden state/.style={context,fill=context color!30},
    update/.style={operation, inner sep=5pt, fill=shared color},
    upsampler/.style={operation, regular polygon, regular polygon sides=4, inner sep=1pt, fill=shared color},
    xcit part/.style={operation, fill=xcit color!70},
    xcit description/.style={description, inner sep=0pt},
    xcit layer/.style={operation, fill=xcit color},
    update tensor/.style={tensor, minimum width=3em+1ex, minimum height=2em, fill=context color!70},
    xcit output/.style={update tensor, fill=xcit color!50},
    motion encoder output/.style={update tensor, fill=shared color!25},
    motion feature/.style={tensor, align=center, fill=shared color!45},
    one scale/.style={rectangle, draw=scale color},
    brace/.style={decorate,decoration={brace,raise=2pt}},
    description/.style={font=\scriptsize},
    image/.style={tensor, fill=image color},
    conv/.style={operation, fill=conv-no-act color},
    conv down/.style={conv, fill=conv color},
    residual unit/.style={operation, fill=res color},
    intermediate feature/.style={feature, fill=intermediate feature color},
    matching block/.style={operation, fill=shared color, minimum height=5.5em, minimum width=4em},
    loop/.style={circle, minimum size=2*\LoopSize, inner sep=0pt, append after command={\pgfextra{\draw[->] (\tikzlastnode.30) arc (30:150:\LoopSize); \draw[->] (\tikzlastnode.210) arc (210:330:\LoopSize);}}},
    smaller font/.style={font=\scriptsize},
    feature projection/.style={operation, align=center, fill=feat proj color},
    key/.style={tensor, fill=kqv color},
    query/.style={tensor, fill=kqv color},
    value/.style={tensor, fill=kqv color},
    cost/.style={tensor, fill=feature color!30},
    res conv/.style={operation, fill=resconv color},
    context res conv/.style={operation, fill=resconv color!60},
    split/.style={operation, fill=cont_act!80},
    output/.style={thick},
}

\def \BelowResOffset {1ex}
\def \DefaultDistance {1em-1.5pt}
\def \InputOffset {2pt}
\def \LoopSize {1em}
\def \LevelDistance {1.5em}
\def \LevelOffset {6em}
\def \AroundXcitDistance {\sequenceDistance+1ex}
\newlength{\initFlowDistance}
\newlength{\innerDistance}
\newlength{\sequenceDistance}
\newlength{\featureDistance}
\newlength{\contextDistance}
\newlength{\initFlowOffset}
\def \AroundXcitDistanceS {\sequenceDistanceS+0.7ex}
\newlength{\initFlowDistanceS}
\newlength{\innerDistanceS}
\newlength{\sequenceDistanceS}
\newlength{\featureDistanceS}
\newlength{\contextDistanceS}
\newlength{\initFlowOffsetS}
\newcommand{\down}{\ensuremath{\downarrow}}
\newcommand{\up}{\ensuremath{\uparrow}}
\newcommand{\Star}{\ensuremath{*}}

\newcommand{\globalcontext}[1]{\ensuremath{\mathit{GC}_{#1}}}
\newcommand{\contextguidedmotionfeature}[1]{\ensuremath{\mathit{CMF}{#1}}}
\newcommand{\motionfeature}[1]{\ensuremath{\mathit{MF}{#1}}}
\newcommand{\context}[1]{\ensuremath{\mathit{C}_{#1}}}
\newcommand{\inputcontext}[1]{\ensuremath{\hat{\mathit{C}}_{#1}}}
\newcommand{\feature}[2]{\ensuremath{\mathit{F}_{#1,#2}}}
\newcommand{\inputfeature}[2]{\ensuremath{\hat{\mathit{F}}_{#1,#2}}}
\newcommand{\hiddenstate}[1]{\ensuremath{\mathit{H}_{#1}}}

%% file: figures/teaser_new.tex
\NewDocumentCommand{\flowimage}{m o}{\begin{tikzpicture}
        \node [anchor=north west, inner sep=0pt] (image) at (0,0) {\includegraphics[width=\textwidth]{#1}};
        \begin{scope}[x={(image.north east)}, y={(image.south west)}]
            \draw [black, densely dotted] (0.82, 0.05) rectangle (0.97, 0.42);
            \draw [black, densely dotted] (0.51, 0.44) rectangle (0.65, 0.76);
            \IfNoValueF{#2}{\node [anchor=south west, font=\tiny, inner sep=1pt] at (0, 1) {#2};}
        \end{scope}
    \end{tikzpicture}}
    \begin{subfigure}{0.49\columnwidth}
        \flowimage{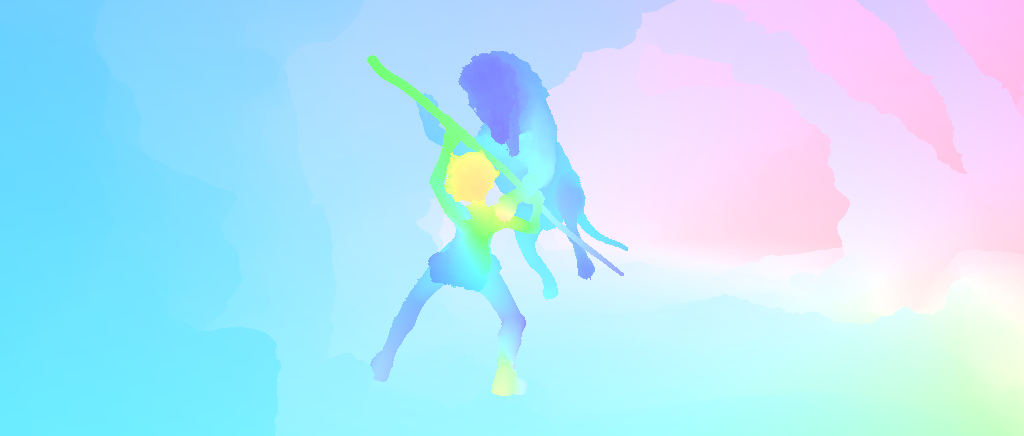}
        
        \vspace{-17.4mm}
        \hspace{0.5mm}\includegraphics[width=0.2\columnwidth]{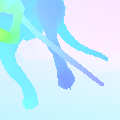}
        
        \hspace{0.5mm}\includegraphics[width=0.2\columnwidth]{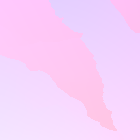}

        \vspace{4.5mm}
        
        \vspace{-5mm}
        \caption{Ground Truth}
    \end{subfigure}
    \begin{subfigure}{0.49\columnwidth}
        \flowimage{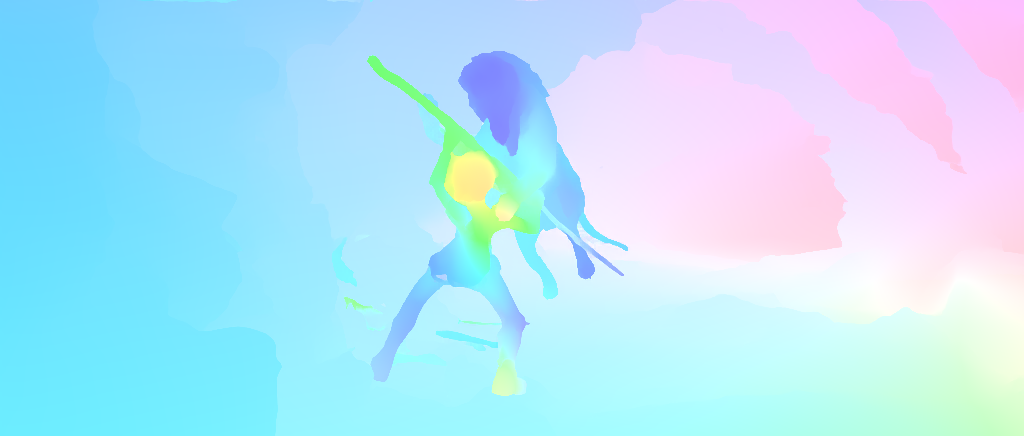}[] 

        \vspace{-17.4mm} 
        \hspace{0.5mm}\includegraphics[width=0.2\columnwidth]{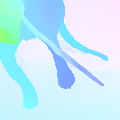}
        
        \hspace{0.5mm}\includegraphics[width=0.2\columnwidth]{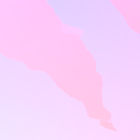}

        \vspace{4.5mm}
        
        \vspace{-5mm}
        \caption{Ours}
    \end{subfigure}
    \\
    \begin{subfigure}{0.49\columnwidth}
        \flowimage{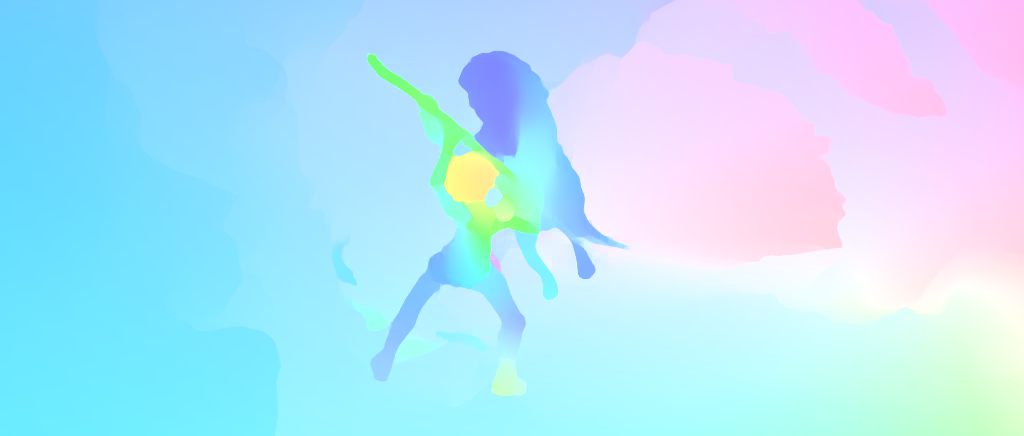}[] 

        \vspace{-17.4mm}
        \hspace{0.5mm}\includegraphics[width=0.2\columnwidth]{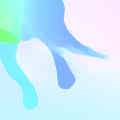}
        
        \hspace{0.5mm}\includegraphics[width=0.2\columnwidth]{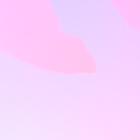}

        \vspace{4.5mm}
        
        \vspace{-5mm}
        \caption{FlowFormer++}
    \end{subfigure}
    \begin{subfigure}{0.49\columnwidth}
        \flowimage{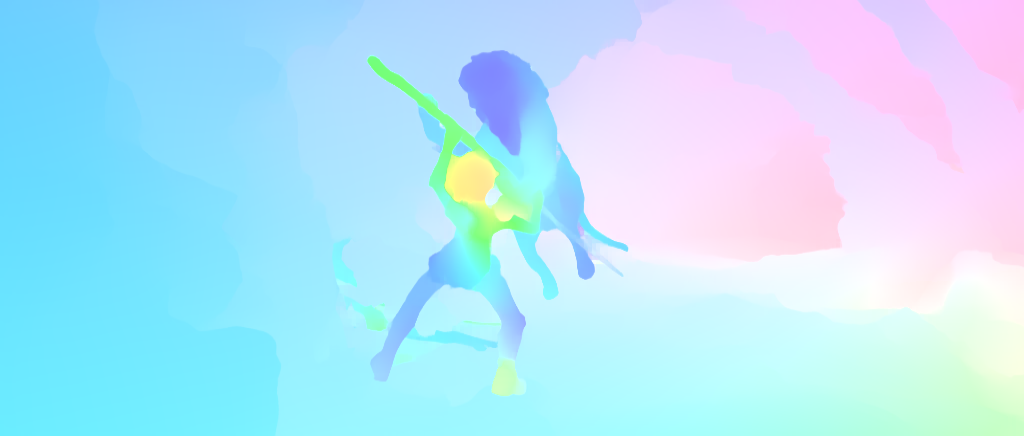}[] 

        \vspace{-17.4mm}
        \hspace{0.5mm}\includegraphics[width=0.2\columnwidth]{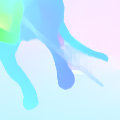}
        
        \hspace{0.5mm}\includegraphics[width=0.2\columnwidth]{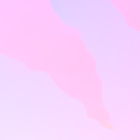}

        \vspace{4.5mm}
        
        \vspace{-5mm}
        \caption{GMFlow+}
    \end{subfigure}
    \\
    \begin{subfigure}{0.49\columnwidth}
        \flowimage{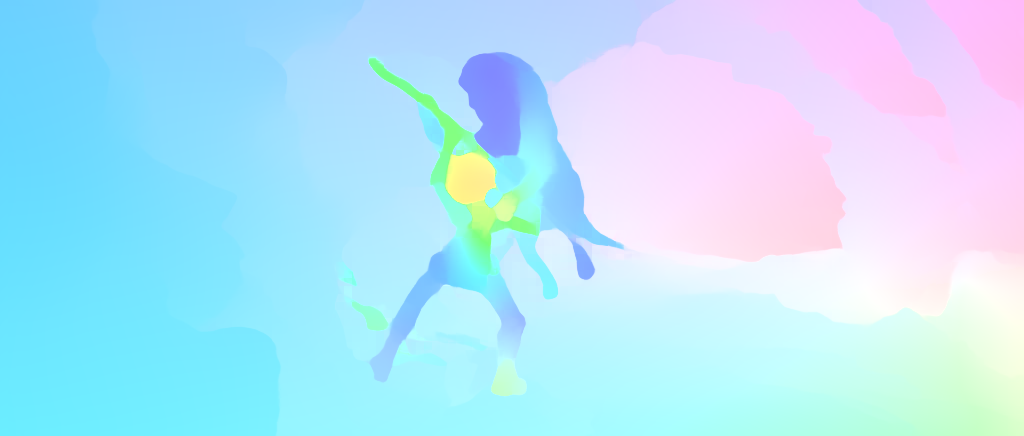}[] 

        \vspace{-17.4mm}
        \hspace{0.5mm}\includegraphics[width=0.2\columnwidth]{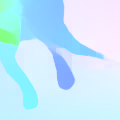}
        
        \hspace{0.5mm}\includegraphics[width=0.2\columnwidth]{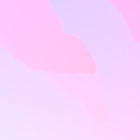}

        \vspace{4.5mm}
        
        \vspace{-5mm}
        \caption{GMA}
    \end{subfigure}
    \begin{subfigure}{0.49\columnwidth}
        \flowimage{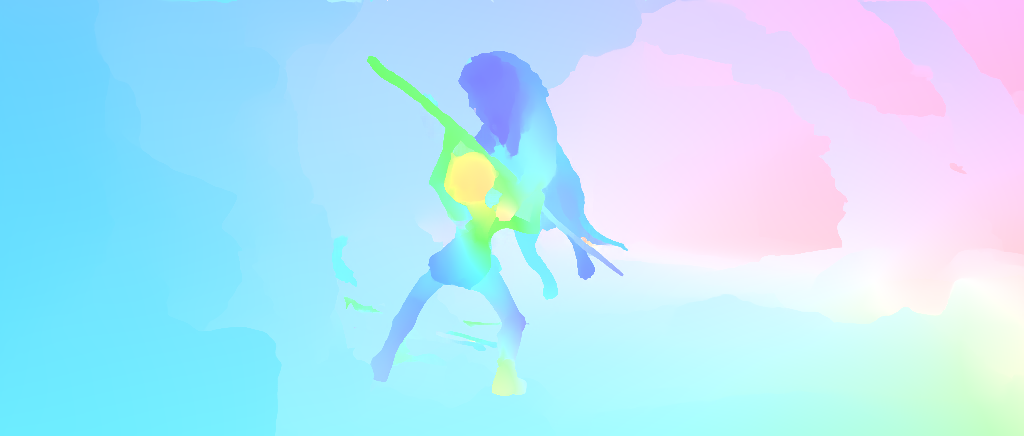}[] 

        \vspace{-17.4mm}
        \hspace{0.5mm}\includegraphics[width=0.2\columnwidth]{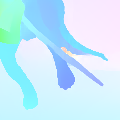}
        
        \hspace{0.5mm}\includegraphics[width=0.2\columnwidth]{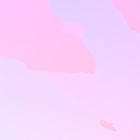}

        \vspace{4.5mm}
        
        \vspace{-5mm}
        \caption{MS-RAFT+}
    \end{subfigure}
   

%% file: figures/multi-scale.tex
\begin{tikzpicture}
    \footnotesize
    \coordinate (hidden state 0);
    \coordinate[yshift=13em] (hidden state 3) at (hidden state 0);
    \foreach \level in {1, 2, 4}
    {
        \pgfmathtruncatemacro{\previousLevel}{\level-1}
        \pgfmathsetmacro{\scaleDenominator}{int(pow(2, 5-\level))}
        \pgfmathsetmacro{\nextScaleDenominator}{int(pow(2, 4-\level))}

        \node [flow, above=\LevelDistance of hidden state \previousLevel.north east, anchor=south east] (flow \level) {\ifnumcomp{\level}{=}{1}{init flow}{flow}};
        \node [feature, above=\InputOffset of flow \level.north east, anchor=south east] (feature \level 2) {\feature{\level}{2}};
        \node [feature, left=\InputOffset of feature \level 2] (feature \level 1) {\feature{\level}{1}};
        \node [initial hidden state, above=\InputOffset of feature \level 2.north east, anchor=south east] (hidden state \level) {\hiddenstate{\level}};
        \node [context, left=\InputOffset of hidden state \level] (context \level) {\context{\level}};
        \node [context, left=\InputOffset of context \level] (global context \level) {\globalcontext{\level}};

        \node [matching block, right=\DefaultDistance of {$(flow \level.east)!0.5!(hidden state \level.east)$}, align=center] (matching block \level) {matching \\ module};
        \node [loop, right=0pt of matching block \level] (loop \level) {$T_{\level}$};
        \node [flow, right=0pt of loop \level] (new flow \level) {flow};
        \node [upsampler, right=\DefaultDistance of new flow \level] (upsampler \level) {\up};
        \node [flow, right=\DefaultDistance of upsampler \level] (upsampled flow \level) {flow};

        \draw[->] (hidden state \level) -- (context \level -| matching block \level.west);
        \draw[->] (feature \level 2) -- (feature \level 2 -| matching block \level.west);
        \draw[->] (flow \level) -- (flow \level -| matching block \level.west);
        \draw[->] (new flow \level) -- (upsampler \level);
        \draw[->] (upsampler \level) -- (upsampled flow \level);

        \draw[brace] (upsampled flow \level.south east) -- (upsampled flow \level.south west) node[description, midway, below, yshift=-2pt] {$\ifthenelse{\nextScaleDenominator=1}{}{\frac{1}{\nextScaleDenominator}{\times}} (h, w)$};
        \draw[brace] (flow \level.south west -| global context \level.west) -- (global context \level.north west) node[description, midway, left, xshift=-2pt] {$\frac{1}{\scaleDenominator}{\times}(h, w)$};
    }

    \foreach \level in {1}
    {
        \pgfmathtruncatemacro{\nextLevel}{\level+1}

        \draw [->] (upsampled flow \level) |- ($(matching block \level)!0.5!(matching block \nextLevel)$) -| (flow \nextLevel);
    }
    \node[xshift=-3em] at ($(flow 4.south -| global context 2.north west)!0.5!(global context 2.north west)$) {\vbox{\baselineskip=0.33333\normalbaselineskip\hbox{\textbf{.}}\hbox{\textbf{.}}\hbox{\textbf{.}}}};
    \draw [->] (upsampled flow 2) |- node[pos=0.25, fill=white] {\vbox{\baselineskip=0.33333\normalbaselineskip\hbox{\textbf{.}}\hbox{\textbf{.}}\hbox{\textbf{.}}}} ($(matching block 2)!0.5!(matching block 4)$) -| (flow 4);
\end{tikzpicture}

%% file: figures/context-feature-transform.tex
\begin{tikzpicture}
    \footnotesize
    \def \level {s}

    \node [intermediate feature] (input feature \level 2) {\inputfeature{\level}{2}};
    \node [feature, anchor=west] at (input feature \level 2.east) (up feature \level-1 2) {\vphantom{\inputfeature{\level}{2}}$\up\feature{\level-1}{2}$};

    \node [intermediate feature, above=\featureDistance of input feature \level 2, anchor=center] (input feature \level 1) {\inputfeature{\level}{1}};
    \node [feature, anchor=west] at (input feature \level 1.east) (up feature \level-1 1) {\vphantom{\inputfeature{\level}{1}}$\up\feature{\level-1}{1}$};

    \node [intermediate context, below=\featureDistance of input feature \level 2.south east, anchor=north east] (input context \level) {\vphantom{$\up\context{\level-1}$}\inputcontext{\level}};
    \node [context, anchor=west] at (input context \level.east) (up context \level-1) {\vphantom{\inputcontext{\level}}$\up\context{\level-1}$};

    \node [res conv, right=\sequenceDistance of up feature \level-1 2] (res conv feature 2 \level) {Res. Conv.$_s$};
    \node [res conv] at (res conv feature 2 \level |- up feature \level-1 1) (res conv feature 1 \level) {Res. Conv.$_s$};
    \node [context res conv] at (res conv feature 1 \level |- up context \level-1) (res conv context \level) {Res. Conv.$_s$};

    \node [split, right=\sequenceDistance of res conv context \level] (split \level) {Split};

    \node [context activation, right=\sequenceDistance of split \level, yshift=0.7ex, anchor=south west]  (tanh \level) {Tanh};
    \node [context activation, below=1.4ex of tanh \level] (relu \level) {ReLU};

    \node [initial hidden state, output, right=\sequenceDistance of tanh \level] (initial hidden state \level) {\hiddenstate{\level}};
    \node [context, output] at (initial hidden state \level |- relu \level) (context \level) {\context{\level}};

    \node [feature, output] at (res conv feature 1 \level -| context \level) (feature \level 1) {\feature{\level}{1}};
    \node [feature, output] at (res conv feature 2 \level -| context \level) (feature \level 2) {\feature{\level}{2}};

    \node [context, output, below=\contextDistance of context \level] (global context \level) {\globalcontext{\level}};
    \node [xcit part, left=\AroundXcitDistance of global context \level] (ffn \level) {FFN};
    \node [xcit part, left=\innerDistance of ffn \level] (lpi \level) {LPI};
    \node [xcit part, left=\innerDistance of lpi \level] (xca \level) {XCA};
    \node [query, left=\innerDistance of xca \level] (query \level) {$Q$};
    \node [feature projection, left=\sequenceDistance of query \level] (feature projection \level) {Feature \\ Proj.};
    \node [key, above=1pt of query \level] (key \level) {$K$};
    \node [value, below=1pt of query \level] (value \level) {$V$};
    \node [xcit description, anchor=south, xshift=3pt] at (value \level.south -| lpi \level) {Self-Attention XCiT$_\level$};
    
    \node [inner sep=0pt, fit=(feature projection \level) (key \level) (value \level) (ffn \level)] (inner xcit layer) {};
    \node [xcit description, anchor=south] at (inner xcit layer.north) (global context computation) {Global Context Computation};
    \begin{scope}[on background layer]
    \node [xcit layer, fit=(inner xcit layer) (global context computation)] {};
    \end{scope}

    \draw[->] (up feature \level-1 2) -- (res conv feature 2 \level);
    \draw[->] (up feature \level-1 1) -- (res conv feature 1 \level);
    \draw[->] (up context \level-1) -- (res conv context \level);

    \draw[->] (res conv feature 2 \level) -- (feature \level 2);
    \draw[->] (res conv feature 1 \level) -- (feature \level 1);

    \draw[->] (res conv context \level) -- (split \level);
    \draw[->] (split \level) |- (tanh \level);
    \draw[->] (tanh \level) -- (initial hidden state \level);
    \draw[->] (split \level) |- (relu \level);
    \draw[->] (relu \level) -- (context \level);
    \draw[->] (context \level.east) -- +(1ex,0) -- +(1ex,-4ex) -| ($(feature projection \level.west) + (-1em, 0)$) -- (feature projection \level);

    \coordinate (border right) at ($(context \level.east) + (1em, -3ex)$);
    \coordinate (border left) at ($(border right -| input feature \level 1.west) - (1em, 0)$);
    \draw [dashed] (border left) -- (border right);

    \draw[->] (feature projection \level) -- (query \level);
    \draw[->] (feature projection \level.340) -- (feature projection \level.340 -| {$(feature projection \level.east)!0.5!(query \level.west)$}) |- (value \level);
    \draw[->] (feature projection \level.20) -- (feature projection \level.20 -| {$(feature projection \level.east)!0.5!(query \level.west)$}) |- (key \level); 
    \draw[->] (query \level) -- (xca \level);
    \draw[->] (key \level) -| (xca \level.140);
    \draw[->] (value \level) -| (xca \level.220);
    \draw[->] (xca \level) -- (lpi \level);
    \draw[->] (lpi \level) -- (ffn \level);
    \draw[->] (ffn \level) -- (global context \level);
    
\end{tikzpicture}

%% file: figures/feat_map_new.tex
\begin{subfigure}[c]{0.95\columnwidth}%
    \footnotesize
    \begin{subfigure}[c]{1.75cm}
    \centering Image $I_1$
    \end{subfigure}
     \begin{subfigure}[c]{0.25\textwidth}
        \includegraphics[width=\textwidth]{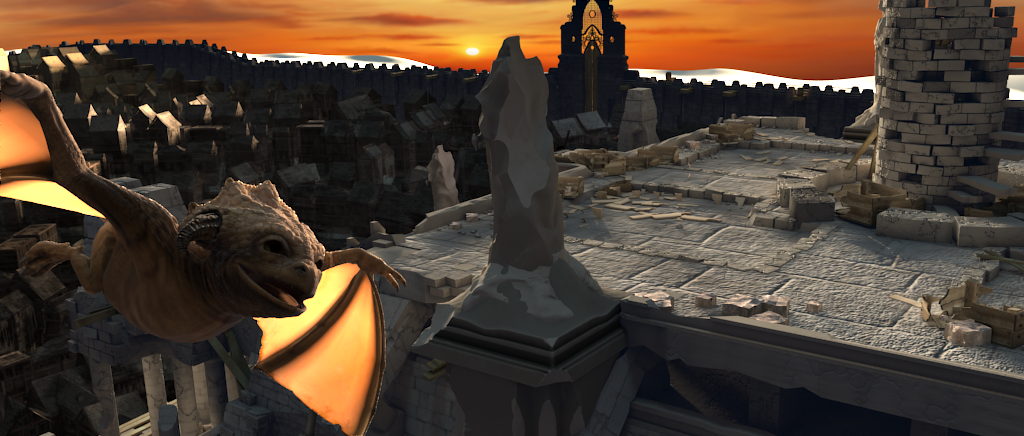}
        \end{subfigure}%
        \hspace{0.5mm}%
    \begin{subfigure}[c]{0.25\textwidth}
            \includegraphics[width=\textwidth]{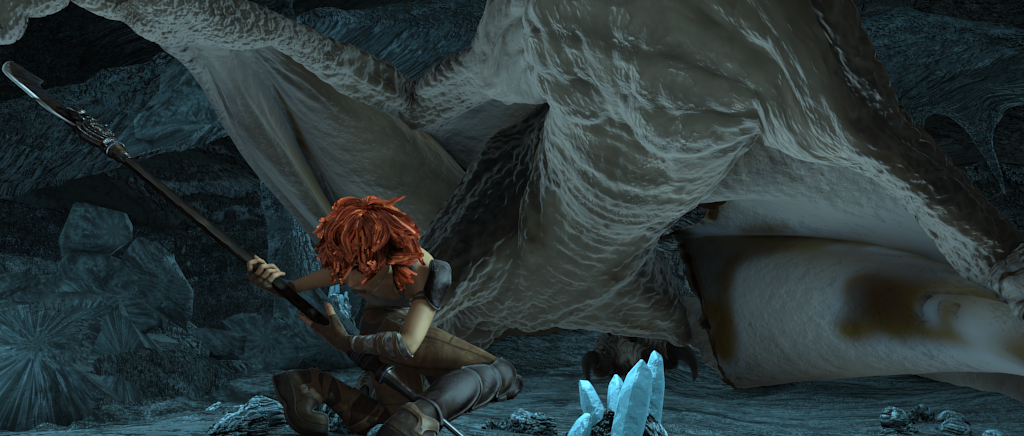}
        \end{subfigure}%
        \hspace{0.5mm}%
         \begin{subfigure}[c]{0.25\textwidth}
            \includegraphics[width=\textwidth]{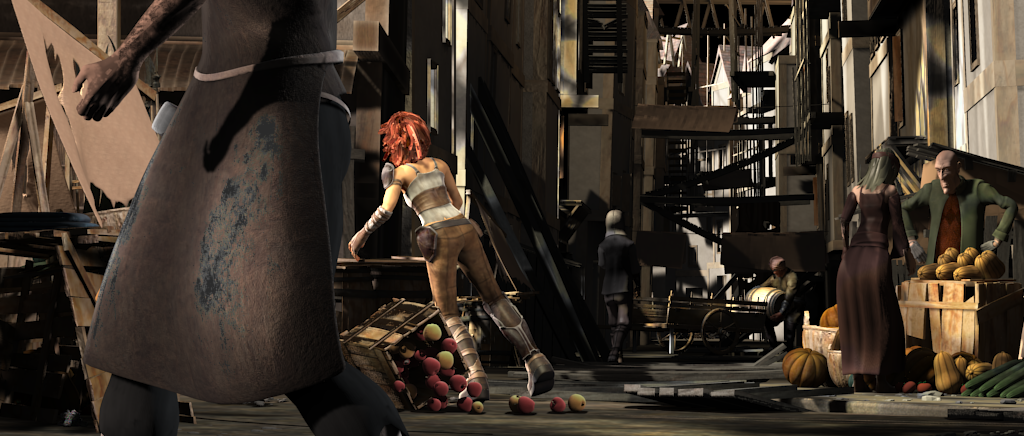}
        \end{subfigure}
        \\
        \begin{subfigure}[c]{1.75cm}
        \centering $\lVert C_{4} \rVert$
        \end{subfigure}   
        \begin{subfigure}[c]{0.25\textwidth}
            \includegraphics[width=\textwidth]{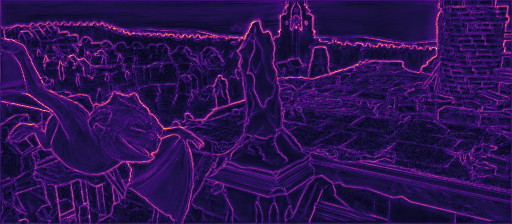}
        \end{subfigure}%
          \hspace{0.5mm}%
        \begin{subfigure}[c]{0.25\textwidth}
            \includegraphics[width=\textwidth]{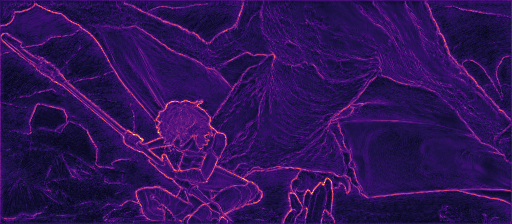}
        \end{subfigure}%
          \hspace{0.5mm}%
         \begin{subfigure}[c]{0.25\textwidth}
            \includegraphics[width=\textwidth]{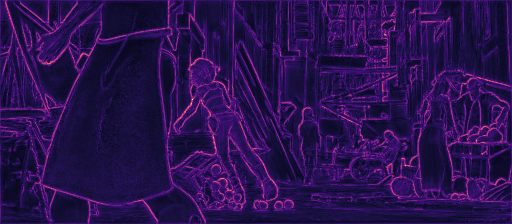}
        \end{subfigure}
        \\
        \begin{subfigure}[c]{1.75cm}
        \centering $\lVert GC_{4} \rVert$
        \end{subfigure}
        \begin{subfigure}[c]{0.25\textwidth}
            \includegraphics[width=\textwidth]{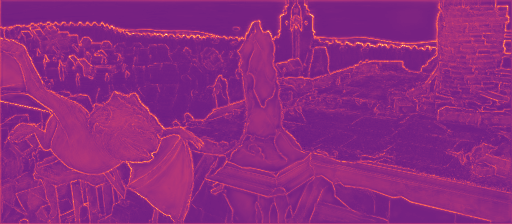}
        \end{subfigure}%
        \hspace{0.5mm}%
        \begin{subfigure}[c]{0.25\textwidth}
            \includegraphics[width=\textwidth]{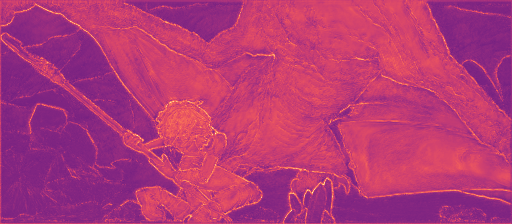}
        \end{subfigure}%
        \hspace{0.5mm}%
        \begin{subfigure}[c]{0.25\textwidth}
            \includegraphics[width=\textwidth]{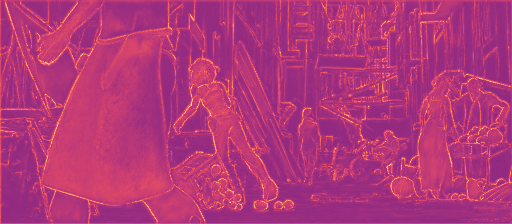}
        \end{subfigure}\\ 
        \begin{subfigure}[c]{1.75cm}
       \centering $\lvert GC_{4}[70] \rvert$
        \end{subfigure}   
        \begin{subfigure}[c]{0.25\textwidth}
            \includegraphics[width=\textwidth]{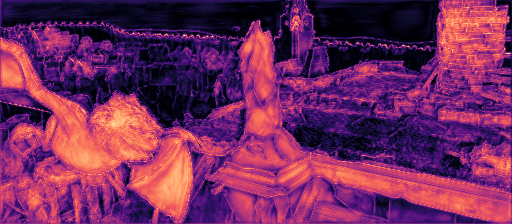}
        \end{subfigure}%
        \hspace{0.5mm}%
        \begin{subfigure}[c]{0.25\textwidth}
            \includegraphics[width=\textwidth]{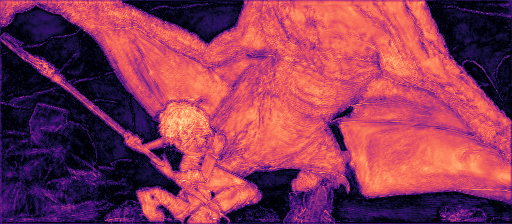}
        \end{subfigure}%
        \hspace{0.5mm}%
        \begin{subfigure}[c]{0.25\textwidth}
            \includegraphics[width=\textwidth]{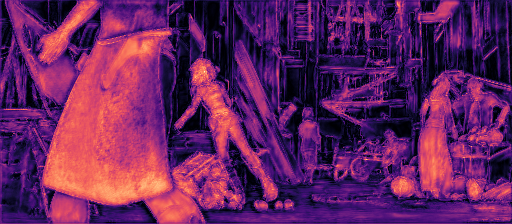}
        \end{subfigure}
        \\
        \begin{subfigure}[c]{1.75cm}
        \centering $\lvert GC_{4}[96] \rvert$
        \end{subfigure}   
     \begin{subfigure}[c]{0.25\textwidth}
            \includegraphics[width=\textwidth]{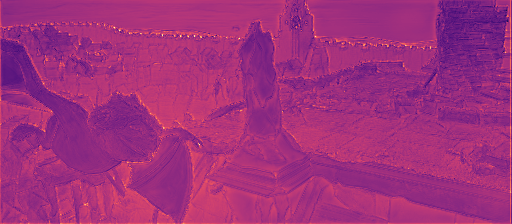}
        \end{subfigure}%
        \hspace{0.5mm}%
        \begin{subfigure}[c]{0.25\textwidth}
            \includegraphics[width=\textwidth]{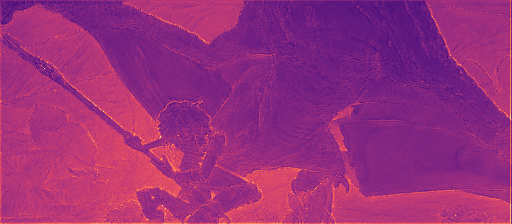}
        \end{subfigure}%
        \hspace{0.5mm}%
        \begin{subfigure}[c]{0.25\textwidth}
            \includegraphics[width=\textwidth]{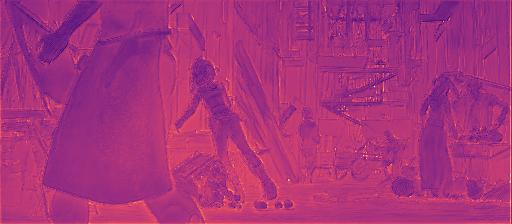}
        \end{subfigure}
    \end{subfigure}%
    %
    \hspace{0.4mm}%
    \begin{subfigure}[c]{0.044\columnwidth}
        \vspace{0pt}
        \includegraphics[width=\textwidth]{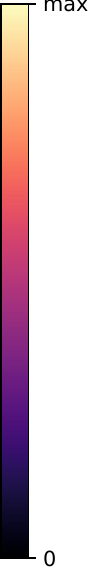}%
    \end{subfigure}   

%% file: figures/matching-block.tex
\begin{tikzpicture}
    \footnotesize
    \def \level {s}

    \node [feature, anchor=center] (feature \level 1) {\feature{\level}{1}};
    \node [feature, above=\featureDistance of feature \level 1, anchor=center] (feature \level 2) {\feature{\level}{2}};
    \node [context, above=\contextDistanceS of feature \level 2, anchor=center] (context \level) {\globalcontext{\level}};

    \node [cost sampling, right=of {$(feature \level 1)!0.5!(feature \level 2)$}, xshift=-10px] (cost sampling \level) {Cost \\ Sampling};
    \node [cost, right=\sequenceDistanceS of cost sampling \level] (cost \level) {Cost};
    \node [flow, below=\initFlowOffset of feature \level 1.south west, anchor=north west] (init flow \level) {Init. Flow};

    \node [motion encoder, right=\sequenceDistanceS of cost \level] (motion encoder \level) {Motion \\ Encoder};
    \node [motion feature, right=\sequenceDistanceS of motion encoder \level] (motion feature \level) {\motionfeature{}};
    \node [feature projection, right=\AroundXcitDistanceS of context \level] (xcit feature projection \level) {Feature\\Proj.$1$};
    \node [feature projection, below=\innerDistance of xcit feature projection \level] (motion feature projection \level) {Feature\\Proj.$2$};
    \node [value, right=\sequenceDistance of motion feature projection \level] (value \level 2) {$V$};

    \node [key, right=\sequenceDistance of xcit feature projection \level] (key \level 2) {$K$};
    \node [query] (query \level 2) at ($(key \level 2)!0.5!(value \level 2)$) {$Q$};

    \node [xcit part, right=\innerDistance of query \level 2] (xca \level 2) {XCA};
    \node [xcit part, right=\innerDistance of xca \level 2] (lpi \level 2) {LPI};
    \node [xcit part, right=\innerDistance of lpi \level 2] (ffn \level 2) {FFN};
    \node [xcit description, anchor=south] at (motion feature projection \level.south -| lpi \level 2) (xcit description \level 2) {Cross-Attention XCiT$_\level$};

    \node [inner sep=0pt, fit=(xcit feature projection \level) (motion feature projection \level) (xca \level 2) (lpi \level 2) (ffn \level 2) (xcit description \level 2)] (inner xcit layer \level 2) {};
    \node [xcit description, anchor=south] at (inner xcit layer \level 2.north) (motion grouping based on global context) {Motion Grouping Based on Global Context};
    \begin{scope}[on background layer]
    \node[xcit layer, fit=(inner xcit layer \level 2) (motion grouping based on global context)] {};
    \end{scope}

    \node [xcit output, right=\AroundXcitDistanceS of ffn \level 2] (xcit output \level) {\contextguidedmotionfeature{}};
    \node [update tensor, anchor=south] at (xcit output \level.north) (xcit output \level top) {\context{\level}};
    \node [motion encoder output, anchor=north] at (xcit output \level.south) (xcit output \level bottom) {\motionfeature{}};

    \node [update, right=\sequenceDistanceS of xcit output \level] (update \level) {Update};
    \node [initial hidden state, above=\innerDistance of update \level] (initial hidden state \level) {\hiddenstate{\level}};

    \node [flow, below=\sequenceDistance of update \level] (flow \level) {Flow};

    \draw[->] (feature \level 1) -- (feature \level 1 -| cost sampling \level.west);
    \draw[->] (feature \level 2) -- (feature \level 2 -| cost sampling \level.west);

    \draw[->] (context \level) -- (xcit feature projection \level);
    \draw[->] (xcit feature projection \level) -- (key \level 2);
    \draw[->] (xcit feature projection \level.340) -- (xcit feature projection \level.340 -| {$(xcit feature projection \level.east)!0.5!(query \level 2.west)$}) |- (query \level 2);
    \draw[->] (query \level 2) -- (xca \level 2);
    \draw[->] (key \level 2) -| (xca \level 2.140);
    \draw[->] (value \level 2) -| (xca \level 2.220);

    \draw[->] (xca \level 2) -- (lpi \level 2);
    \draw[->] (lpi \level 2) -- (ffn \level 2);
    \draw[->] (cost sampling \level) -- (cost \level);
    \draw[->] (cost \level) -- (motion encoder \level);
    \draw[->] (motion encoder \level) -- (motion feature \level);
    \draw[->] (motion feature \level.45 |- {$(motion feature \level) + (0, 2em+1ex)$}) -| ($(motion feature projection \level.west) + (-1em, 0)$) -- (motion feature projection \level);
    \draw[->] (motion feature \level.45) |- (xcit output \level bottom);
    \draw[->] (motion feature projection \level) -- (value \level 2);

    \draw[->] (ffn \level 2) -| ($(ffn \level 2.east)!0.5!(ffn \level 2.east -| xcit output \level.west)$) |- (xcit output \level.west);

    \draw[->] (xcit output \level) -- (update \level);
    \draw[->] (initial hidden state \level) -- (update \level);
    \draw[->] (update \level) -- (flow \level);

    \draw[->] (init flow \level.20) -- (init flow \level.20 |- cost sampling \level.south);
    \draw[->] ($(init flow \level.20) + (0, 1ex)$) -| (motion encoder \level);
    \draw[->] (flow \level) |- (init flow \level);

\end{tikzpicture}

%% file: figures/examples_new.tex
\begin{figure*}
    \NewDocumentCommand{\flowimage}{m o}{\begin{tikzpicture}
        \node [anchor=north west, inner sep=0pt] (image) at (0,0) {\includegraphics[width=\textwidth]{#1}};
        \begin{scope}[x={(image.north east)}, y={(image.south west)}]
            \draw [black, densely dotted] (0.01, 0.01) rectangle (0.28, 0.45);
            \draw [black, densely dotted] (0.32, 0.32) rectangle (0.37, 0.5);
            \draw [black, densely dotted] (0.90, 0.59) rectangle (0.99, 0.77);
            \draw [black, densely dotted] (0.35, 0.01) rectangle (0.44, 0.15);
            \IfNoValueF{#2}{\node [anchor=south west, font=\tiny, inner sep=1pt] at (0, 1) {#2};}
        \end{scope}
    \end{tikzpicture}}
    \begin{subfigure}{0.16\textwidth}
        \flowimage{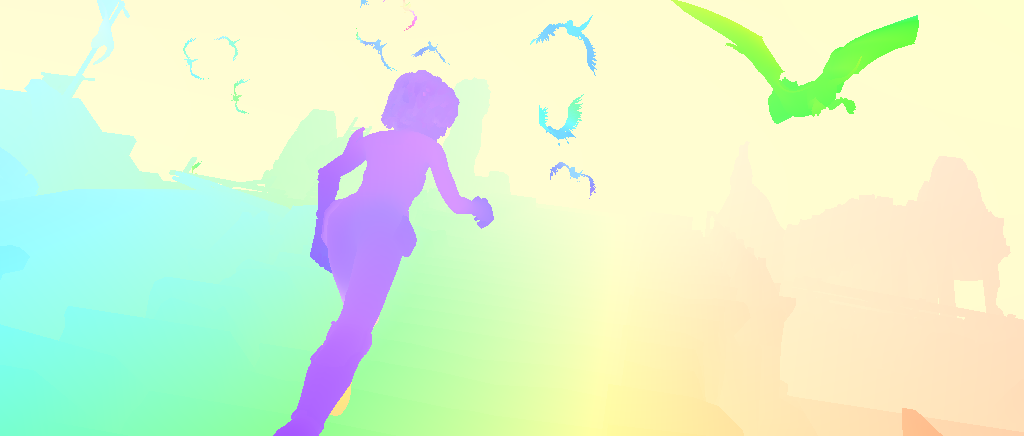}
    \end{subfigure}
    \begin{subfigure}{0.16\textwidth}
        \flowimage{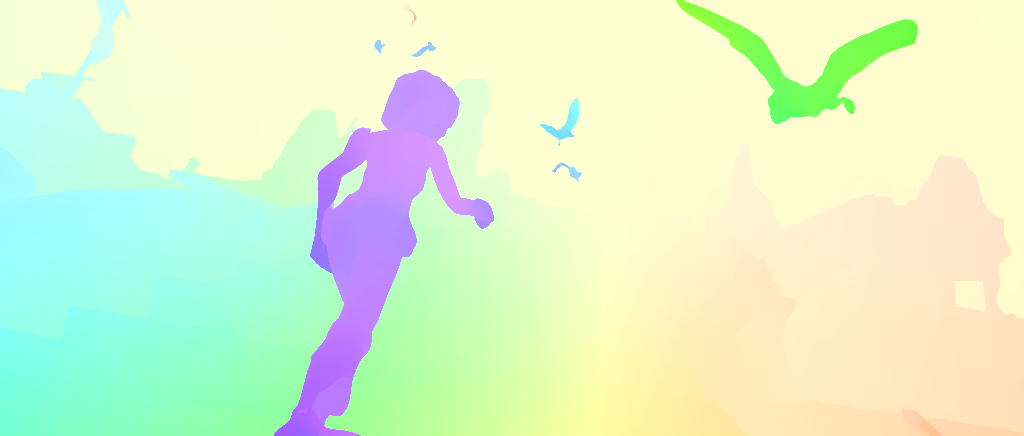}[EPE 0.441]
    \end{subfigure}
    \begin{subfigure}{0.16\textwidth}
        \flowimage{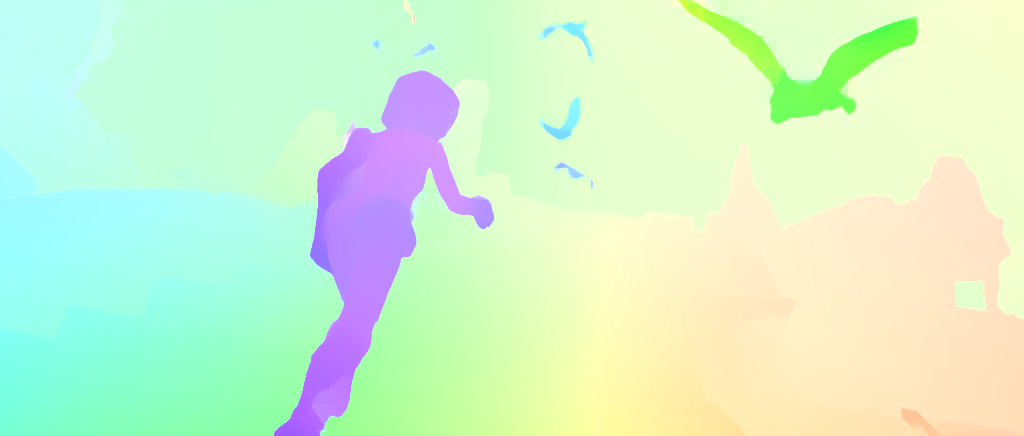}[EPE 0.657]
    \end{subfigure}
    \begin{subfigure}{0.16\textwidth}
        \flowimage{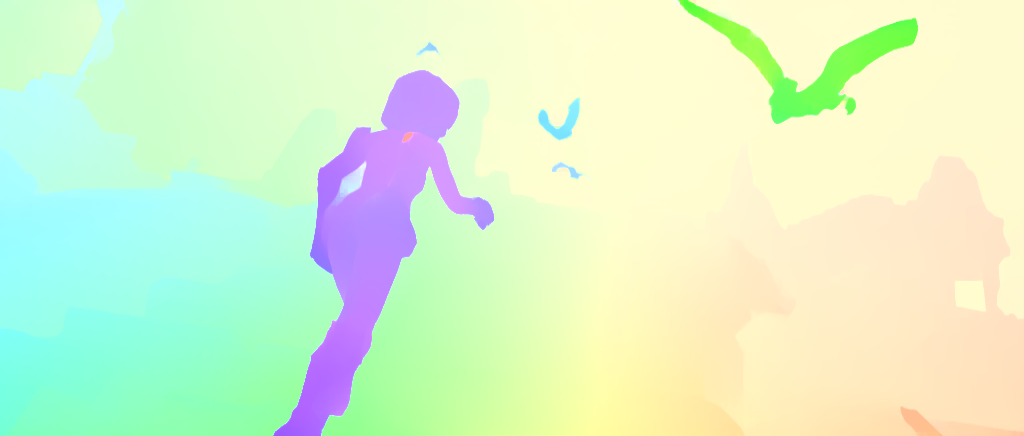}[EPE 0.559]
    \end{subfigure}
    \begin{subfigure}{0.16\textwidth}
        \flowimage{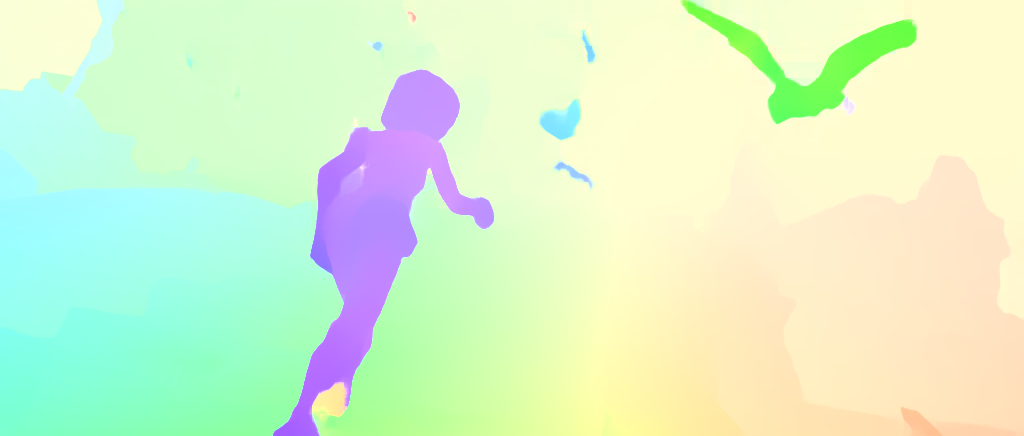}[EPE 0.691]
    \end{subfigure}
    \begin{subfigure}{0.16\textwidth}
        \flowimage{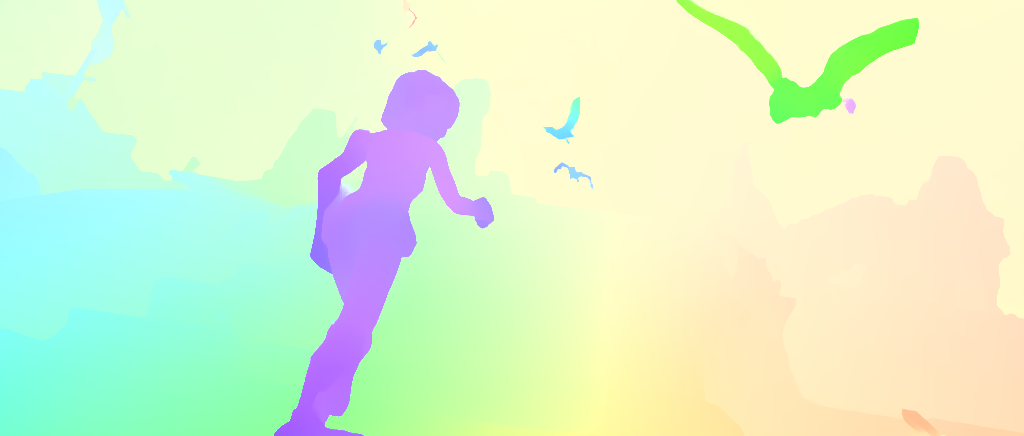}[EPE 0.552]
    \end{subfigure}
    \\[2pt]
    \RenewDocumentCommand{\flowimage}{m o}{\begin{tikzpicture}
        \node [anchor=north west, inner sep=0pt] (image) at (0,0) {\includegraphics[width=\textwidth]{#1}};
        \begin{scope}[x={(image.north east)}, y={(image.south west)}]
            \draw [black, densely dotted] (0.24, 0.48) rectangle (0.27, 0.59);
            \draw [black, densely dotted] (0.37, 0.01) rectangle (0.57, 0.42);
            \draw [black, densely dotted] (0.43, 0.44) rectangle (0.67, 0.76);
            \IfNoValueF{#2}{\node [anchor=south west, font=\tiny, inner sep=1pt] at (0, 1) {#2};}
        \end{scope}
    \end{tikzpicture}}
    \begin{subfigure}{0.16\textwidth}
        \flowimage{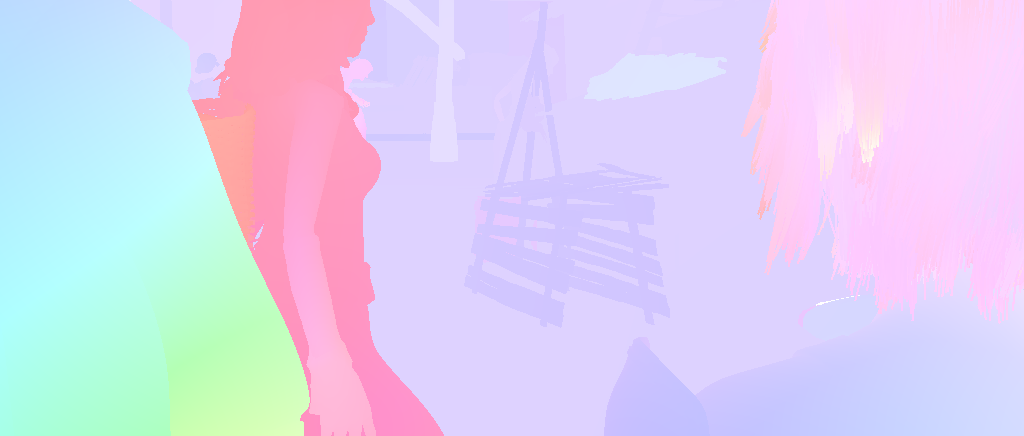}
    \end{subfigure}
    \begin{subfigure}{0.16\textwidth}
        \flowimage{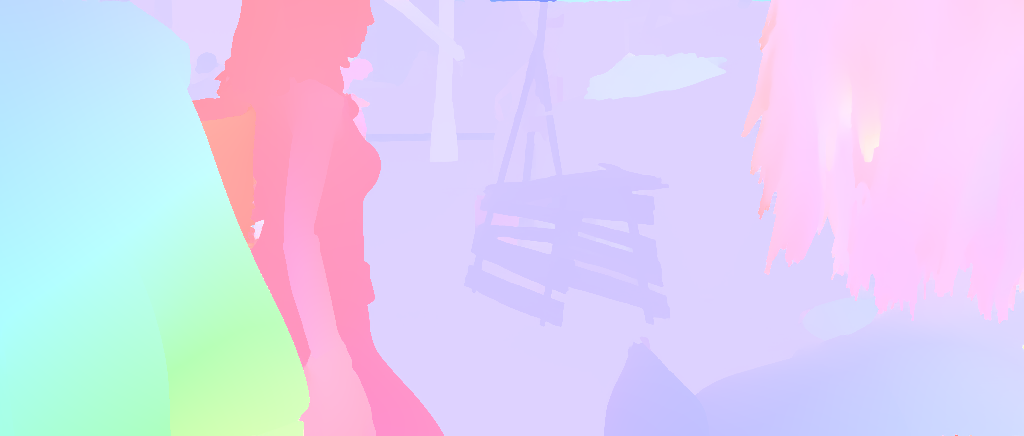}[EPE 0.371]
    \end{subfigure}
    \begin{subfigure}{0.16\textwidth}
        \flowimage{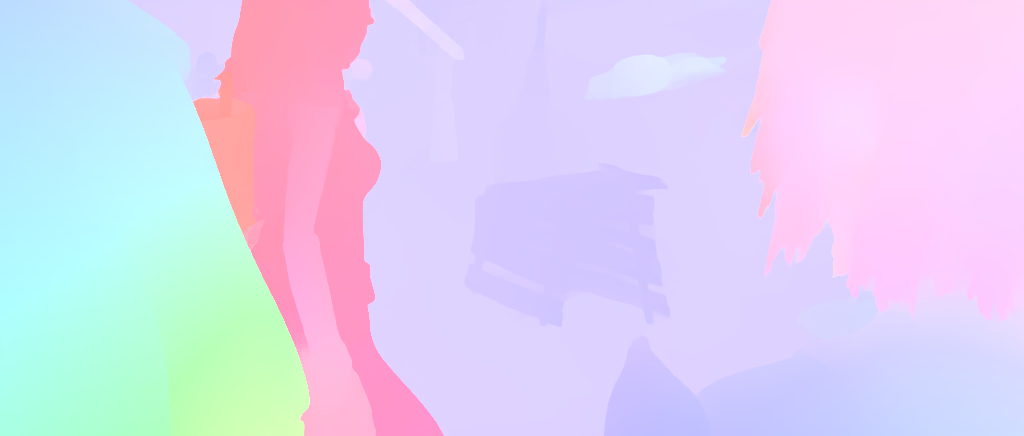}[EPE 0.511]
    \end{subfigure}
    \begin{subfigure}{0.16\textwidth}
        \flowimage{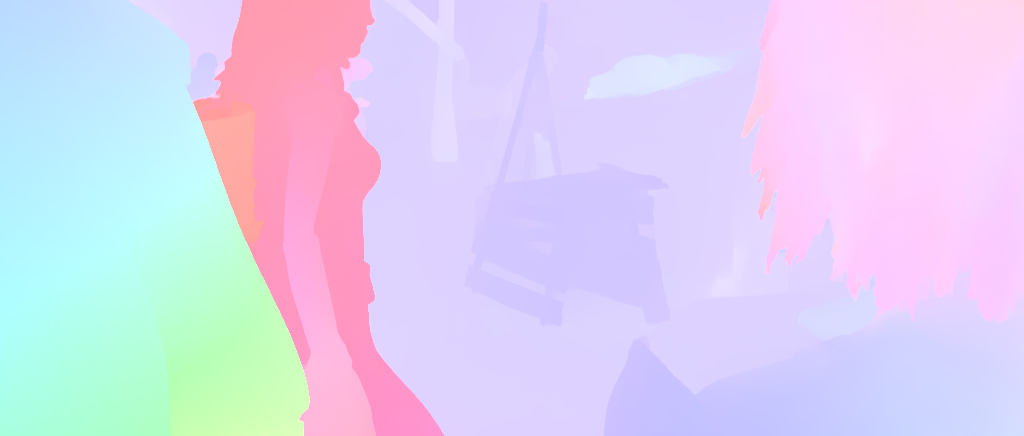}[EPE 0.406]
    \end{subfigure}
    \begin{subfigure}{0.16\textwidth}
        \flowimage{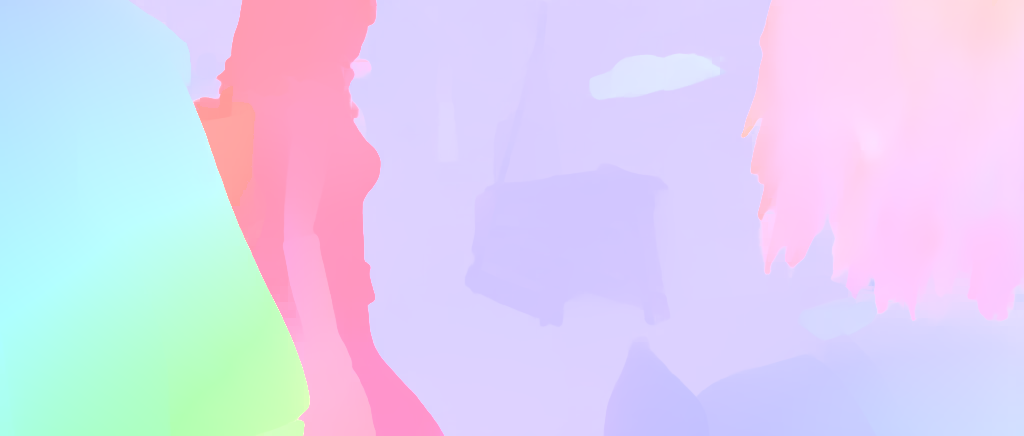}[EPE 0.624]
    \end{subfigure}
    \begin{subfigure}{0.16\textwidth}
        \flowimage{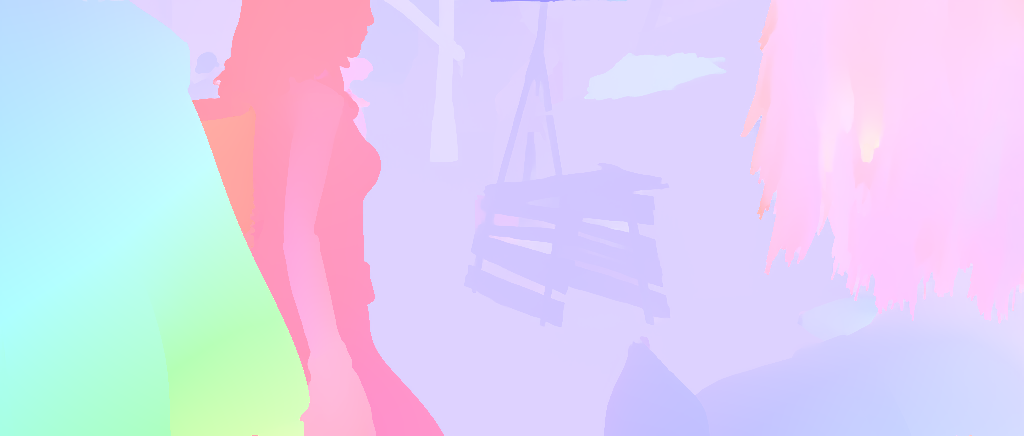}[EPE 0.417]
    \end{subfigure}
    \\[-1pt]
    \begin{subfigure}{0.16\textwidth}
        \caption{Ground Truth}
    \end{subfigure}
    \begin{subfigure}{0.16\textwidth}
        \caption{CCMR+ (Ours)}
    \end{subfigure}
    \begin{subfigure}{0.16\textwidth}
        \caption{FlowFormer++}
    \end{subfigure}
    \begin{subfigure}{0.16\textwidth}
        \caption{GMFlow+}
    \end{subfigure}
    \begin{subfigure}{0.16\textwidth}
        \caption{GMA}
    \end{subfigure}
    \begin{subfigure}{0.16\textwidth}
        \caption{MS-RAFT+}
    \end{subfigure}
    \\[1ex]
    \RenewDocumentCommand{\flowimage}{m o}{\begin{tikzpicture}
        \node [anchor=north west, inner sep=0pt] (image) at (0,0) {\includegraphics[width=\textwidth]{#1}};
        \begin{scope}[x={(image.north east)}, y={(image.south west)}]
            \draw [white, densely dotted] (0.80, 0.35) rectangle (0.97, 0.7);
            \IfNoValueF{#2}{\node [anchor=south west, text=white, font=\tiny, inner sep=1pt] at (0, 1) {#2};}
        \end{scope}
    \end{tikzpicture}}%
    \begin{subfigure}{0.16\textwidth}
        \flowimage{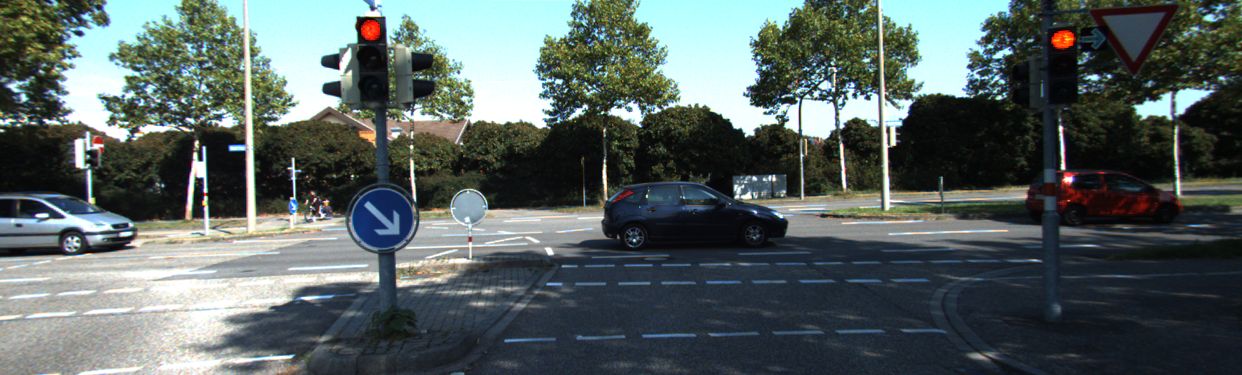}
    \end{subfigure}
    \begin{subfigure}{0.16\textwidth}
        \flowimage{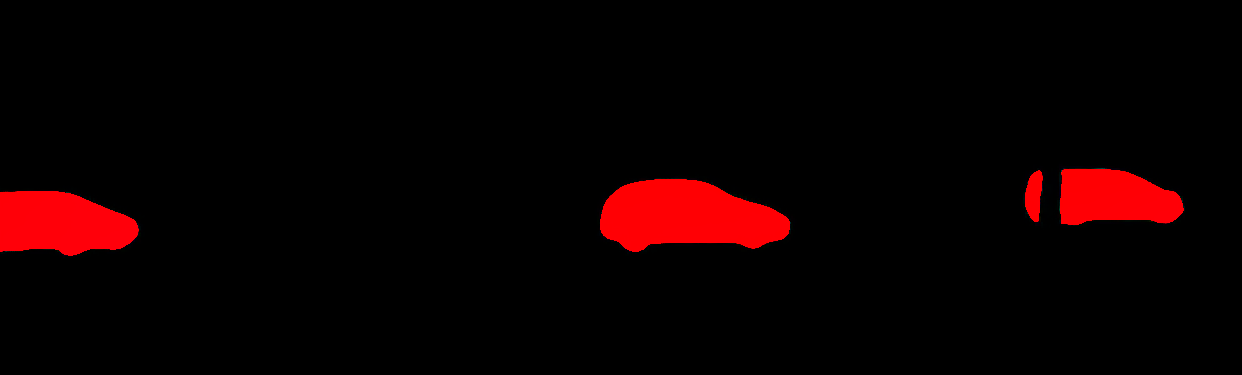}[Fl-all 0.67]
    \end{subfigure}
    \begin{subfigure}{0.16\textwidth}
        \flowimage{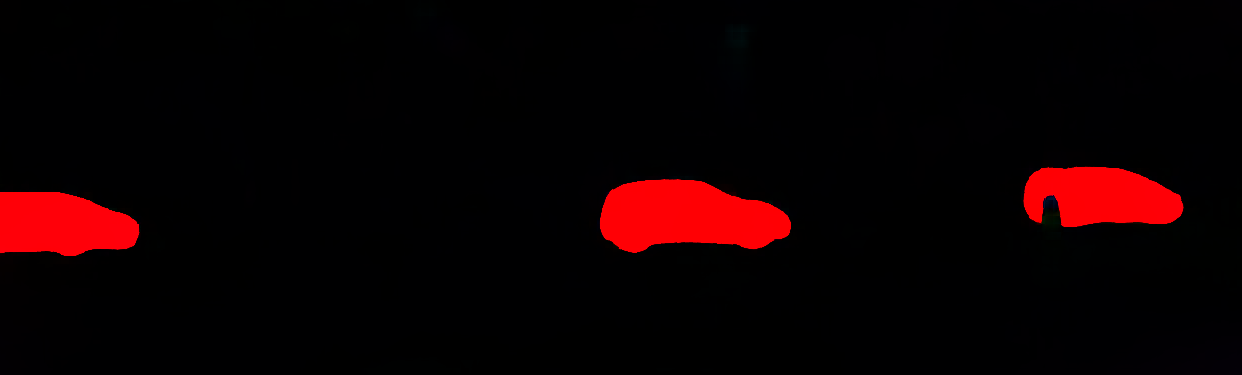}[Fl-all 1.03]
    \end{subfigure}
    \begin{subfigure}{0.16\textwidth}
        \flowimage{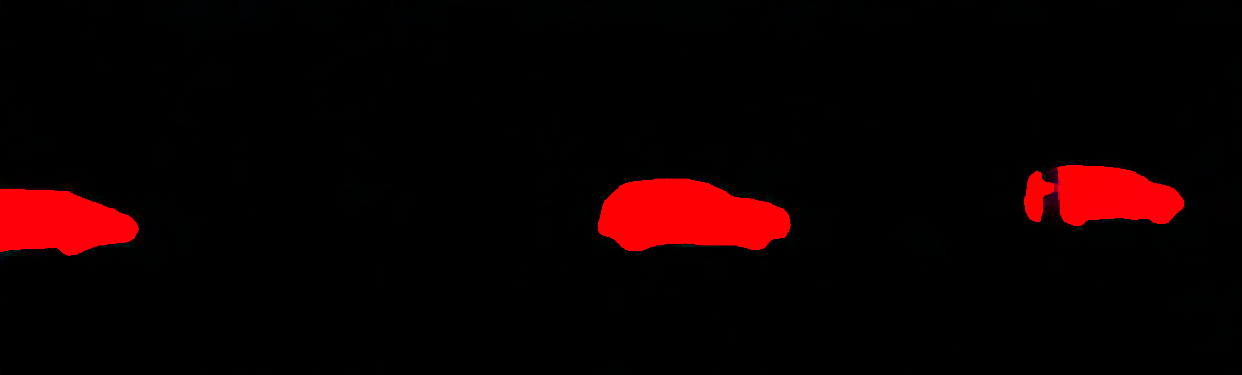}[Fl-all 1.01]
    \end{subfigure}
    \begin{subfigure}{0.16\textwidth}
        \flowimage{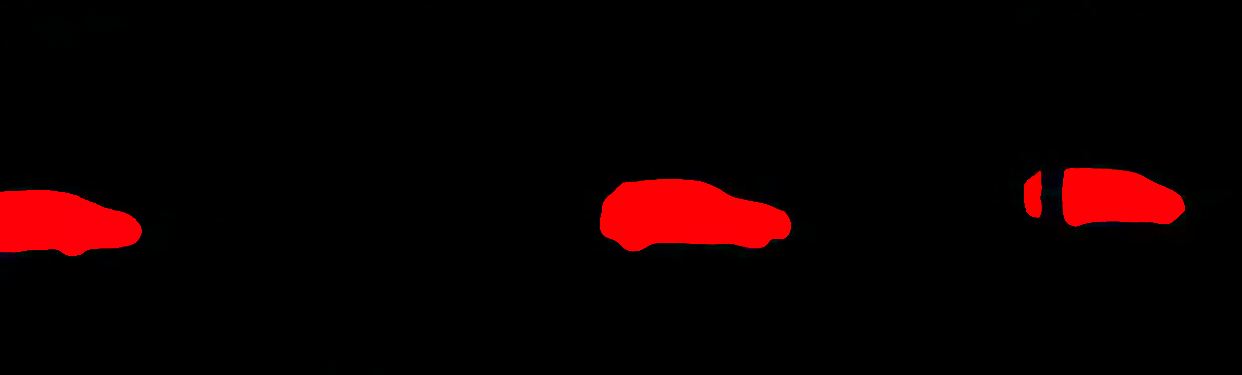}[Fl-all 0.79]
    \end{subfigure}
    \begin{subfigure}{0.16\textwidth}
        \flowimage{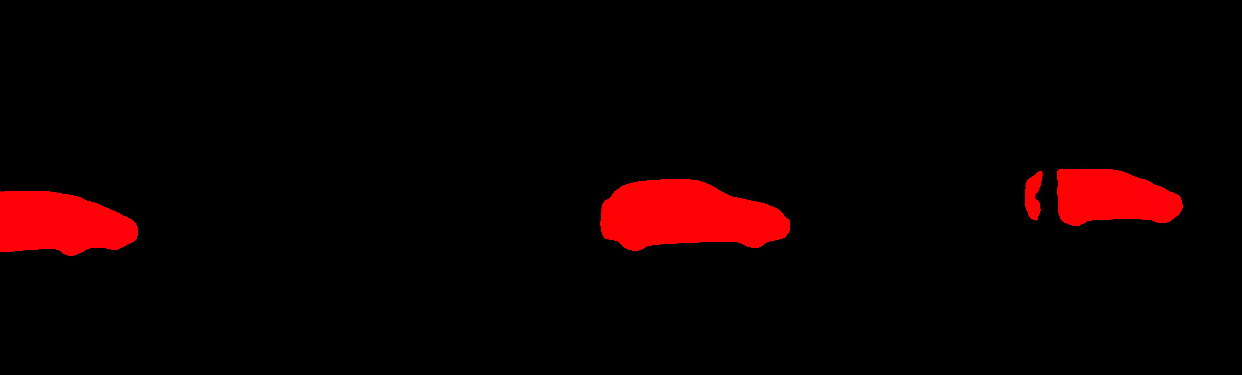}[Fl-all 0.84]
    \end{subfigure}
    \\[2pt]
    \RenewDocumentCommand{\flowimage}{m o}{\begin{tikzpicture}
        \node [anchor=north west, inner sep=0pt] (image) at (0,0) {\includegraphics[width=\textwidth]{#1}};
        \begin{scope}[x={(image.north east)}, y={(image.south west)}]
            \draw [white, densely dotted] (0.28, 0.38) rectangle (0.43, 0.78);
            \IfNoValueF{#2}{\node [anchor=south west, font=\tiny, inner sep=1pt] at (0, 1) {#2};}
        \end{scope}
    \end{tikzpicture}}%
    \begin{subfigure}{0.16\textwidth}
        \flowimage{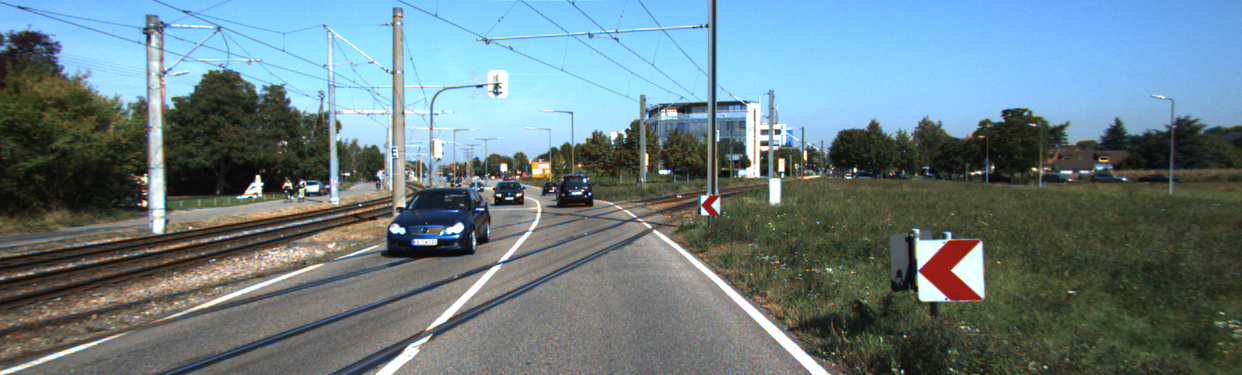}

        \vspace{-7.8mm}
        \hspace{18.9mm}\includegraphics[width=0.3\columnwidth]{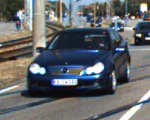}
        
        \vspace{0.7mm}
        
    \end{subfigure}
    \begin{subfigure}{0.16\textwidth}
        \flowimage{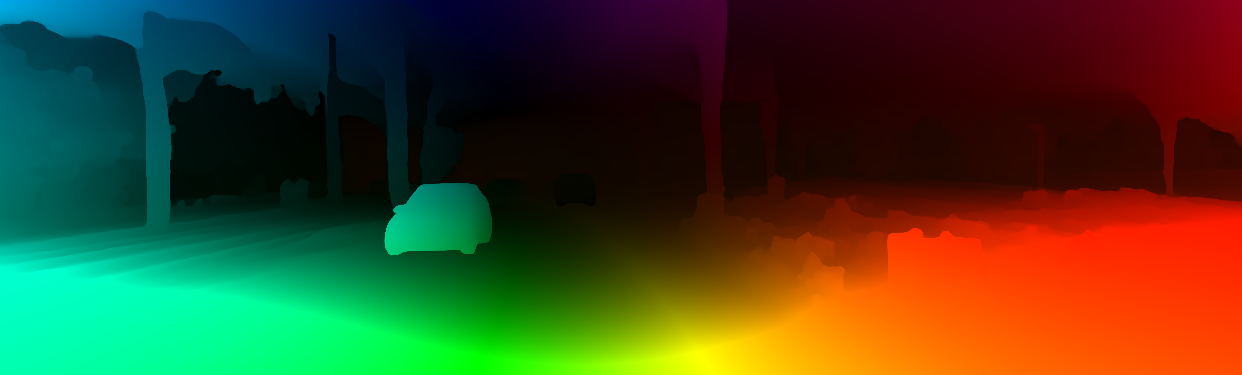}[Fl-all 6.78]
        
        \vspace{-7.8mm}
        \hspace{18.9mm}\includegraphics[width=0.3\columnwidth]{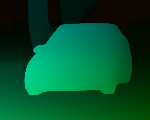}
        
        \vspace{0.7mm}
        
    \end{subfigure}
    \begin{subfigure}{0.16\textwidth}
        \flowimage{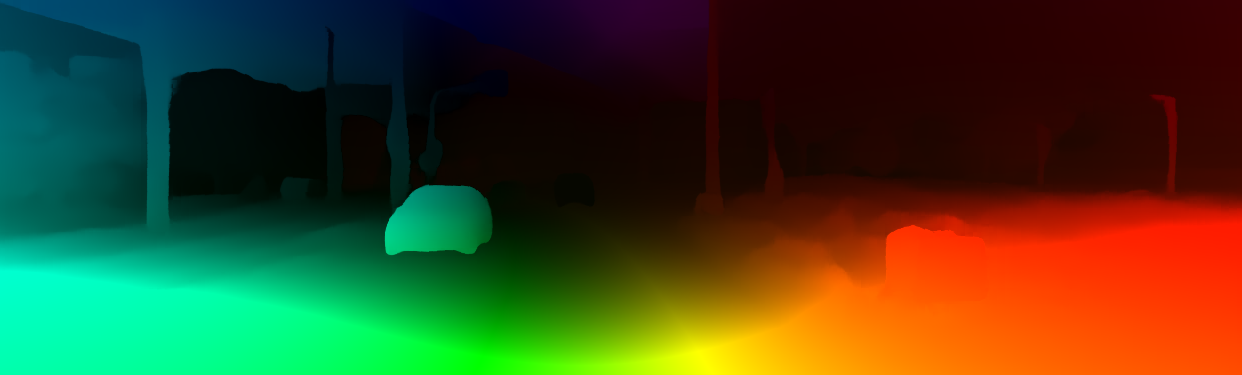}[Fl-all 8.12]

        \vspace{-7.8mm}
        \hspace{18.9mm}\includegraphics[width=0.3\columnwidth]{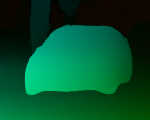}
        
        \vspace{0.7mm}
        
    \end{subfigure}
    \begin{subfigure}{0.16\textwidth}
        \flowimage{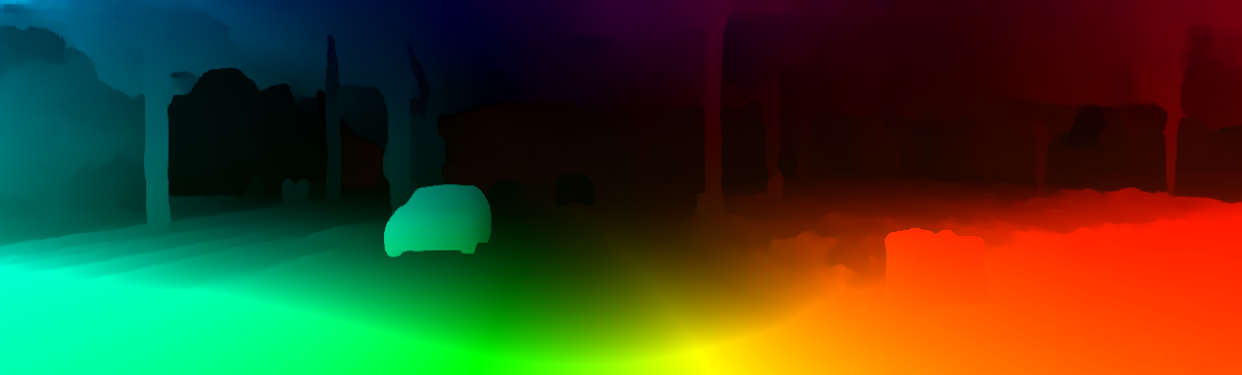}[Fl-all 7.32]

        \vspace{-7.8mm}
        \hspace{18.9mm}\includegraphics[width=0.3\columnwidth]{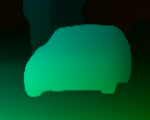}
        
        \vspace{0.7mm}
        
    \end{subfigure}
    \begin{subfigure}{0.16\textwidth}
        \flowimage{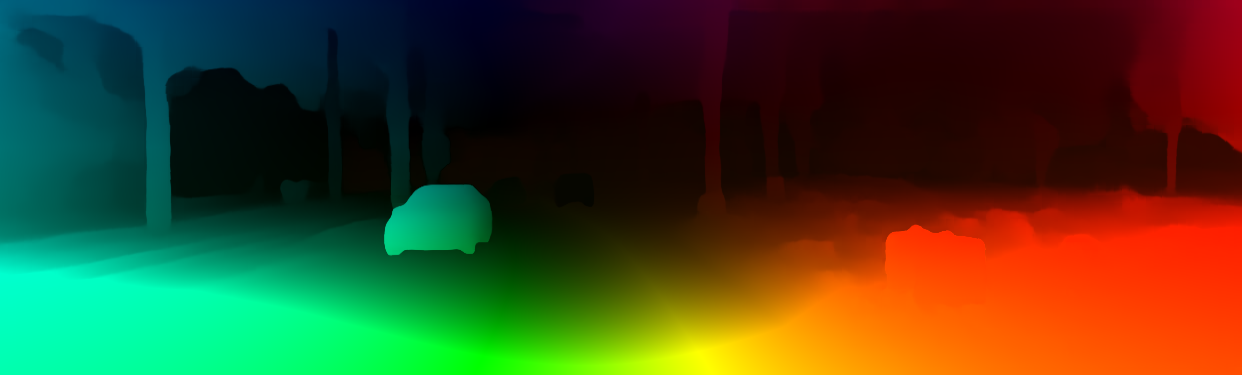}[Fl-all 8.70]

        \vspace{-7.8mm}
        \hspace{18.9mm}\includegraphics[width=0.3\columnwidth]{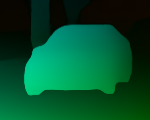}
        
        \vspace{0.7mm}
        
    \end{subfigure}
    \begin{subfigure}{0.16\textwidth}
        \flowimage{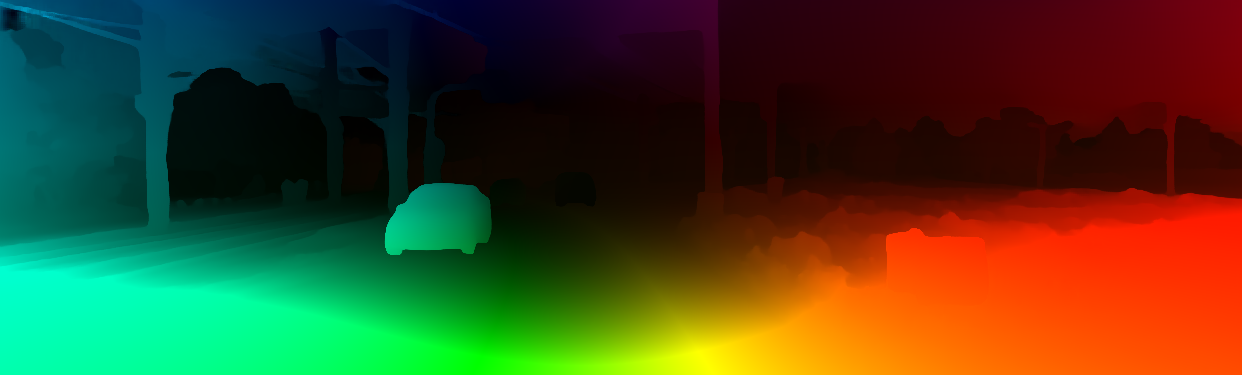}[Fl-all 6.83]

        \vspace{-7.8mm}
        \hspace{18.9mm}\includegraphics[width=0.3\columnwidth]{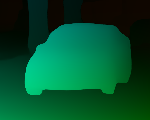}
        
        \vspace{0.7mm}
        
    \end{subfigure}
    \\[-1pt]
    \begin{subfigure}{0.16\textwidth}
         \caption{Image 1}
     \end{subfigure}
     \begin{subfigure}{0.16\textwidth}
         \caption{CCMR+ (Ours)}
     \end{subfigure}
     \begin{subfigure}{0.16\textwidth}
         \caption{AnyFlow}
     \end{subfigure}
     \begin{subfigure}{0.16\textwidth}
         \caption{GMFlow+}
     \end{subfigure}
     \begin{subfigure}{0.16\textwidth}
         \caption{MatchFlow\_GMA}
     \end{subfigure}
     \begin{subfigure}{0.16\textwidth}
         \caption{MS-RAFT+}
     \end{subfigure}
     \vspace{-2mm}
    \caption{Visual comparison of the results of our CCMR+ approach to two recent SOTA methods for each benchmark
    and to the two baselines MS-RAFT+ and GMA, when the original publication results available,
    From top to bottom: Temple-1 (Sintel-final), PERTURBED-market-3 (Sintel-clean), Sequences \#9 and \#15 (KITTI). Details best viewed electronically (by zooming in the flow fields).}
    \label{fig:comparison}
\end{figure*}

%% file: figures/encoder.tex
\begin{tikzpicture}
    \node [image] (image) {Image};
    \node [conv down, below=\DefaultDistance of image] (conv down) {Conv\down};
    \node [context activation, below=\DefaultDistance of conv down] (relu) {ReLU};
    \node [residual unit, below=\DefaultDistance of relu] (res 4) {Res.\vphantom{\down}$_4$};
    \node [residual unit, below=of res 4] (res 3) {Res.\down$_3$};
    \node [residual unit, below=of res 3] (res 2) {Res.\down$_2$};
    \node [residual unit, below=of res 2] (res 1) {Res.\down$_1$};

    \foreach \i in {2, 3, 4}
    {
        \node [intermediate feature, right=2*\DefaultDistance of res \i.south, yshift=-\BelowResOffset] (intermediate feature \i) {$\hat{\mathit{F}}_{\i}$};
        \node [intermediate feature, right=\DefaultDistance of intermediate feature \i] (intermediate feature \i part) {};
        \node [feature, right=0pt of intermediate feature \i part] (upsampled intermediate feature \i part) {};
        \node [res conv, right=\DefaultDistance of upsampled intermediate feature \i part] (new res \i) {Res. Conv.$_\i$};
        \node [feature, right=\DefaultDistance of new res \i] (feature \i) {$\mathit{F}_{\i}$};

        \draw[->] (res \i) |- (intermediate feature \i);
        \draw[->] (intermediate feature \i) -- (intermediate feature \i part);
        \draw[->] (upsampled intermediate feature \i part) -- (new res \i);
        \draw[->] (new res \i) -- (feature \i);

        \node[fit=(intermediate feature \i part) (new res \i), draw, dashed] {};
    }
    
    \node [conv, fill=res color, yshift=-\BelowResOffset] at (res 1.south -| new res 2)  (new res 1) {Conv.};
    \node [feature] at (feature 2 |- new res 1) (feature 1) {$\mathit{F}_{1}$};

    \draw[->] (image) -- (conv down);
    \draw[->] (conv down) -- (relu);
    \draw[->] (relu) -- (res 4);
    \draw[->] (res 4) -- (res 3);
    \draw[->] (res 3) -- (res 2);
    \draw[->] (res 2) -- (res 1);
    \draw[->] (res 1) |- (new res 1);
    \draw[->] (new res 1) -- (feature 1);

    \foreach \i/\j in {1/2, 2/3, 3/4}
    {
        \node[upsampler, yshift=-4pt] at ($(new res \i)!0.5!(new res \j)$) (upsampler \i \j) {\up};
        \draw[->] (feature \i) |- (upsampler \i \j) -| (upsampled intermediate feature \j part);
    }
    \foreach \i in {1, ..., 4}
    {
        \pgfmathsetmacro{\scaleDenominator}{int(pow(2, 5-\i))}
        \draw[brace] (feature \i.south east) -- (feature \i.south west) node[description, midway, below, yshift=-2pt] {$\frac{1}{\scaleDenominator} {\times} (h, w)$};
    }
\end{tikzpicture}